\documentclass[10pt,twocolumn,letterpaper]{article}

\usepackage[pagenumbers]{cvpr} 

\usepackage[dvipsnames]{xcolor}

\usepackage{graphicx}
\usepackage{amsmath}
\usepackage{amsthm}
\usepackage{amssymb}
\usepackage{booktabs}
\usepackage{pifont}
\usepackage[dvipsnames]{xcolor}
\usepackage{enumitem}
\usepackage{pgfplots}
\pgfplotsset{compat=1.18}

\newcommand{\cmark}{\ding{51}}%
\newcommand{\xmark}{\ding{55}}%

\theoremstyle{definition}
\newtheorem{definition}{Definition}[section]

\definecolor{cvprblue}{rgb}{0.21,0.49,0.74}
\usepackage[pagebackref,breaklinks,colorlinks,citecolor=cvprblue]{hyperref}

\usepackage[capitalize]{cleveref}
\crefname{section}{Sec.}{Secs.}
\Crefname{section}{Section}{Sections}
\Crefname{table}{Table}{Tables}
\crefname{table}{Tab.}{Tabs.}
\crefname{definition}{Def.}{Defs.}

\def\paperID{14967} 
\def\confName{CVPR}
\def\confYear{2024}

\title{Improving 2D Human Pose Estimation in Rare Camera Views with Synthetic Data}

\author{\large Miroslav Purkrabek \hspace{0.05cm} and \hspace{0.05cm} Jiri Matas
\vspace{0.12cm}\\
\normalsize
Visual Recognition Group\\
\normalsize
Department of Cybernetics\\
\normalsize
Faculty of Electrical Engineering\\
\normalsize
Czech Technical University in Prague\\
{\tt\small purkrmir@fel.cvut.cz}
}

\begin{document}
\maketitle
\begin{abstract}

Methods and datasets for human pose estimation focus predominantly on side- and front-view scenarios.
%
We overcome the limitation by leveraging synthetic data and 
%
%
introduce RePoGen (RarE POses GENerator), an SMPL-based method for generating synthetic humans with comprehensive control over pose and view.
%
Experiments on top-view datasets and a new dataset of real images with diverse poses show that adding the RePoGen data to the COCO dataset outperforms previous approaches to top- and bottom-view pose estimation without harming performance on common views.
%
An ablation study shows that anatomical plausibility, a property prior research focused on, is not a prerequisite for effective performance. 
%
The introduced dataset and the corresponding code are available on the \hyperlink{https://mirapurkrabek.github.io/RePoGen-paper/}{project website}\footnote{\url{https://mirapurkrabek.github.io/RePoGen-paper/}}.

\end{abstract}    
\section{Introduction}
\label{sec:intro}


The availability of large-scale, manually annotated datasets has greatly advanced research in human pose estimation from 2D monocular images which is closely linked to applications like gesture recognition and action recognition.
Current datasets (eg. \cite{COCO, MPII, LSP}) dominantly contain images from what we call \textit{an orbital view}, i.e. side, front, and back views, where challenges such as occlusion by objects or individuals are most important. 
They focus on everyday activities like standing, sitting, and walking.
Therefore, much of the research has been dedicated to addressing occlusion and specialized datasets (\cite{OCHuman, CrowdPose}) have been curated to evaluate the effectiveness of pose estimation models in scenarios involving occluded individuals.

The issue of unusual viewpoints has received less attention.
In what we refer to as \textit{extreme viewpoints} (top and bottom view; the complement of orbital view), the appearance of humans differs significantly from that of the orbital view.
Although such views are less common in everyday scenarios, they are important for action, activity and gesture recognition in sports and surveillance videos, particularly during transitions between two orbital views.
Annotating persons in extreme views poses considerable challenges, as human annotators struggle to comprehend scenes unfamiliar to the human eye.

\begin{figure}[tb]
  \centering
   \includegraphics[width=\linewidth]{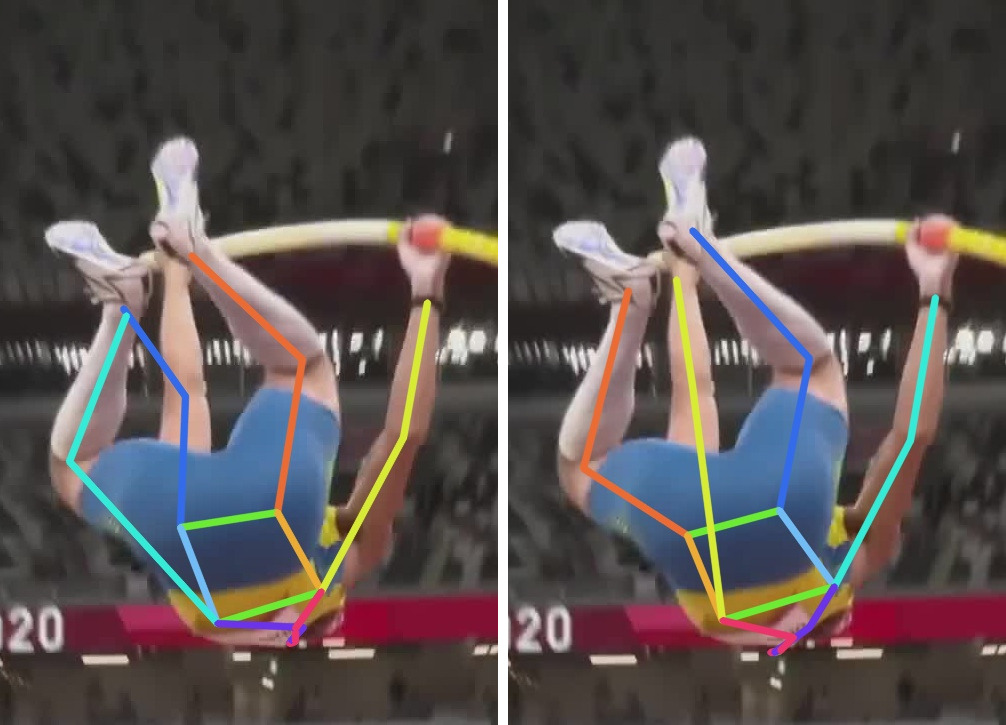}

   \caption{
   Pose estimation trained on COCO (left) and 
   by our method (right).
   The COCO trained model swaps the left and right sides and interprets the \colorbox{yellow}{right hand} as the \colorbox{blue}{\textcolor{white}{left leg}} and the \colorbox{orange}{right leg} as the \colorbox{SkyBlue}{left hand} (color codes the corresponding label).}
   \label{fig:intro}
\end{figure}

We propose an SMPL%
\footnote{SMPL, Skinned Multi-Person Linear Model is a realistic 3D model of the human body.
\label{footnote}
}
-based \cite{SMPL-X} synthetic data approach similar to \cite{SURREAL} and \cite{PanopTOP} to address the scarcity of training data.
The distinguishing feature of our method is that we permit 
generating novel poses, even if they occasionally deviate from anatomical accuracy. In particular, we allow the possibility that body parts, eg limbs, intersect each other as long as the pose maintains approximated physical plausibility. Minor mesh intersections can simulate body deformations without impeding training.
The approach allows to generate new poses from a wider distribution than previous methods. We show that pose variability, combined with novel views, is crucial for accurate pose estimation in sports, where extreme poses and extreme views are neither rare not irrelevant.

%

In summary, the main contributions of the paper are:

\begin{enumerate}
\item \textbf{RePoGen - a new method for generating} synthetic realistic \textbf{images} of humans in previously unseen poses without the need for expensive 3D scans.

\item \textbf{The RePoGen dataset of synthetic images} prioritizing rare poses and viewpoints.

\item \textbf{Manually annotated RePo dataset of real images} of rare poses from the top and bottom views, enabling comprehensive evaluation of pose estimation from unusual views.


\item The fine-tuned ViTPose \cite{ViTPose} \textbf{model, which improves extreme view pose estimation without compromising COCO performance} by augmenting the COCO dataset with synthetic data from RePoGen.

\end{enumerate}

We will release the RePoGen code, synthetic RePoGen, and real-world annotated RePo datasets. 
Additionally, we provide improved annotations for the previously published PoseFES dataset \cite{PoseFES}.
\section{Related Work}
\label{sec:related}

Numerous datasets have been developed to support progress in human pose estimation. Real-world datasets like COCO \cite{COCO} and MPII \cite{MPII} offer diverse images that capture human poses in everyday scenes, while the LSP dataset \cite{LSP} focuses on sports-related poses. To address the challenge of occlusion, specialized datasets such as OCHuman \cite{OCHuman} and CrowdPose \cite{CrowdPose} have been created.


Several models have emerged, demonstrating that the problems has attracted attention.
These models fall primarily into the top-down class and rely on bounding boxes as input for pose estimation. Among these models, ViTPose \cite{ViTPose} stands out as the current SOTA on the COCO dataset, leveraging the transformer architecture. Models such as SWIN \cite{SWIN} and PSA \cite{PSA} also employ transformer-based architectures, but perform slightly worse than  ViTPose in terms of accuracy.

An alternative approach that has garnered attention is the HRNet model \cite{HRNet}, which combines convolutional neural networks with the so called Unbiased Data Processing \cite{UDP}. This combination yields excellent results and has become a common baseline for evaluating the performance of new pose estimation methods. 

Addressing the challenges posed by occlusion and crowded scenes, specialized models have been developed.
For example, I2RNet \cite{I2RNet} is a transformer-based network designed to tackle the challenges of occlusion and crowd-related issues. 

Furthermore, appropriate data processing techniques have been proposed to enhance the performance of pose estimation models. The DARK algorithm \cite{DARK} and the UDP method (Unbiased Data Processing) \cite{UDP} are two notable papers that highlight the importance of data processing in achieving superior results.

To facilitate the research and development of pose estimation, the MMPose framework \cite{MMPose} has emerged as a comprehensive resource. It offers an extensive model zoo and many pre-trained models, including the widely used HRNet. 

Synthetic datasets have also played an important role in augmenting the available data and expanding the range of pose variations. The THEODORE+ dataset \cite{PoseFES} provides a synthetic collection of top-view videos generated using a game engine. These videos show individuals walking in a room, although they only use 13 keypoints instead of the more commonly used 17. Synthetic datasets like SURREAL \cite{SURREAL} and PanopTOP \cite{PanopTOP} utilize the SMPL model \cite{SMPL}, fitting it to measured 3D point clouds of real poses from datasets such as Human36M \cite{h36m} and Panoptic \cite{Panoptic}. However, PanopTOP has limitations regarding low resolution and issues with ghost hands, which should be considered.

The estimation of poses from extreme viewpoints is another research area of interest. The WEPDTOF-Pose dataset \cite{WEPDTOF-Pose} represents the largest dataset of top-view images for pose estimation. Although specialized for top-view poses, it is noteworthy that most people captured in the dataset are from the orbital view due to fisheye lens distortion. Similarly, the PoseFES dataset \cite{PoseFES}, designed for evaluating top-view human pose estimation, also suffers from a prevalence of orbital views caused by fisheye lens distortion. Another dataset, ITop \cite{ITop}, focuses on pose estimation from top-view depthmaps with no RGB images available.

Data augmentation is critical in addressing the scarcity of annotated real-world data for human pose estimation. Various methods have been introduced to tackle this challenge, often involving human parsing techniques for body part segmentation. HumanPaste \cite{HumanPaste} and AdversarialAugmentation \cite{AdversarialAugmentation} employ strategies to simulate occlusion by pasting additional people or selective body parts. Similarly, JointlyOD \cite{JointlyOD} and NearbyPersonOD \cite{NearbyPersonOD} augment data by introducing body parts or whole bodies to mimic occlusion and crowded scenarios. Hwang et al. \cite{RarePoses} explore the problem of rare poses in 2D.

While these augmentation methods prove effective for specific challenges, they do not directly address the problem of unseen viewpoints. In contrast, generating synthetic data using game engines has been explored to introduce variability. However, datasets created with game engines, such as PoseFES \cite{PoseFES}, PeopleSansPeople \cite{PeopleSansPeople}, or LetsPF \cite{LetsPF}, often suffer from limited pose variability, typically showcasing walking or a narrow range of everyday activities.

Another avenue for synthetic data generation involves fitting the SMPL model \cite{SMPL} to 3D point clouds obtained from motion capture systems. For example, SURREAL \cite{SURREAL} fits the SMPL model to the Human36M dataset, providing a pool of textures applicable to SMPL models. Similarly, PanopTOP \cite{PanopTOP} employs the SMPL model fitted to the Panoptic dataset. However, these methods face challenges in fitting the model to point clouds, resulting in issues such as ghost hands. Furthermore, the limitations of motion capture systems make capturing extreme dynamic poses or new poses challenging. AMASS \cite{AMASS} and AGORA \cite{AGORA} offer a big database of 3D poses from 3D scans.
SyntheticHF \cite{SyntheticHF} estimates the SMPL pose and shape from a monocular image and modifies the shape while preserving the pose, creating data resembling SURREAL and Panoptic. However, this approach has limitations due to the initial SMPL estimate, resulting in difficulties handling poses beyond its accurate capture.

Efforts have also been made to enhance the realism of the SMPL model. SMPL-X \cite{SMPL-X} enhances the previous model with hand poses and facial expressions. PoseNDF \cite{PoseNDF} learns a manifold of known poses, enabling the generation of random realistic poses within the manifold. Similarly, CAPE \cite{CAPE} introduces a clothing layer on top of existing SMPL models, aiming to narrow the domain gap between generated and real data.

GAN-based methods like SynthetizeAnyone \cite{SynthetizeAnyone}, UnpairedPG \cite{UnpairedPG}, and SynthetizingIO \cite{SynthetizingIO} generate synthetic data by preserving the given pose or style. On the other hand, diffusion-based methods such as StableDiffusion \cite{StableDiffusion} and ControlNet \cite{ControlNet} offer promising approaches for synthetic data generation, allowing control over the rendered images. However, both approaches have limitations regarding extreme views and rare poses due to the need for more training data.

\begin{figure}[t]
  \centering
  \begin{subfigure}{\linewidth}
    \centering
    \includegraphics[width=\linewidth]{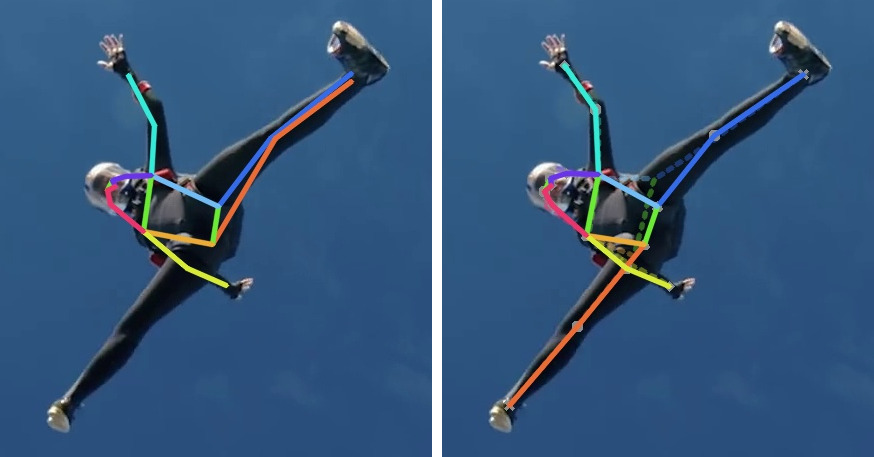}
  \end{subfigure}

  \begin{subfigure}{\linewidth}
    \centering
    \includegraphics[width=\linewidth]{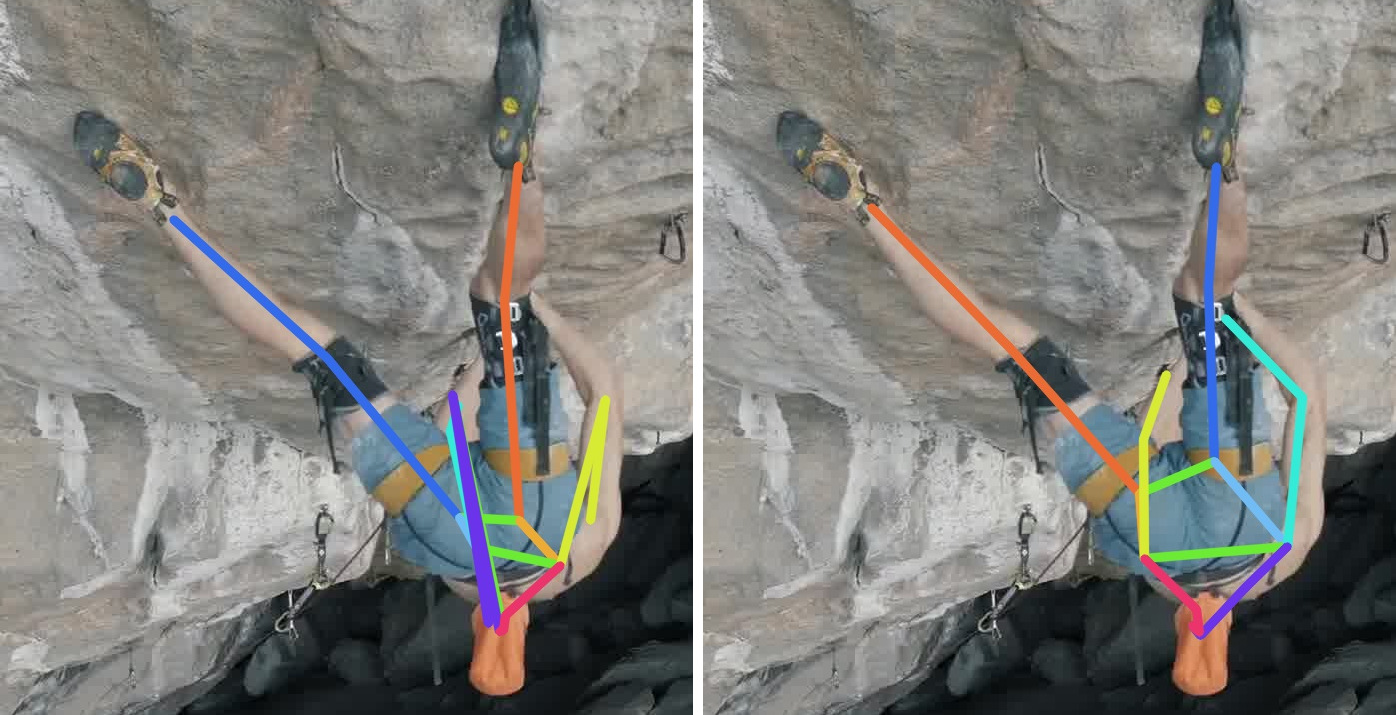}
  \end{subfigure}
    
  \begin{subfigure}{\linewidth}
    \centering
    \includegraphics[width=\linewidth]{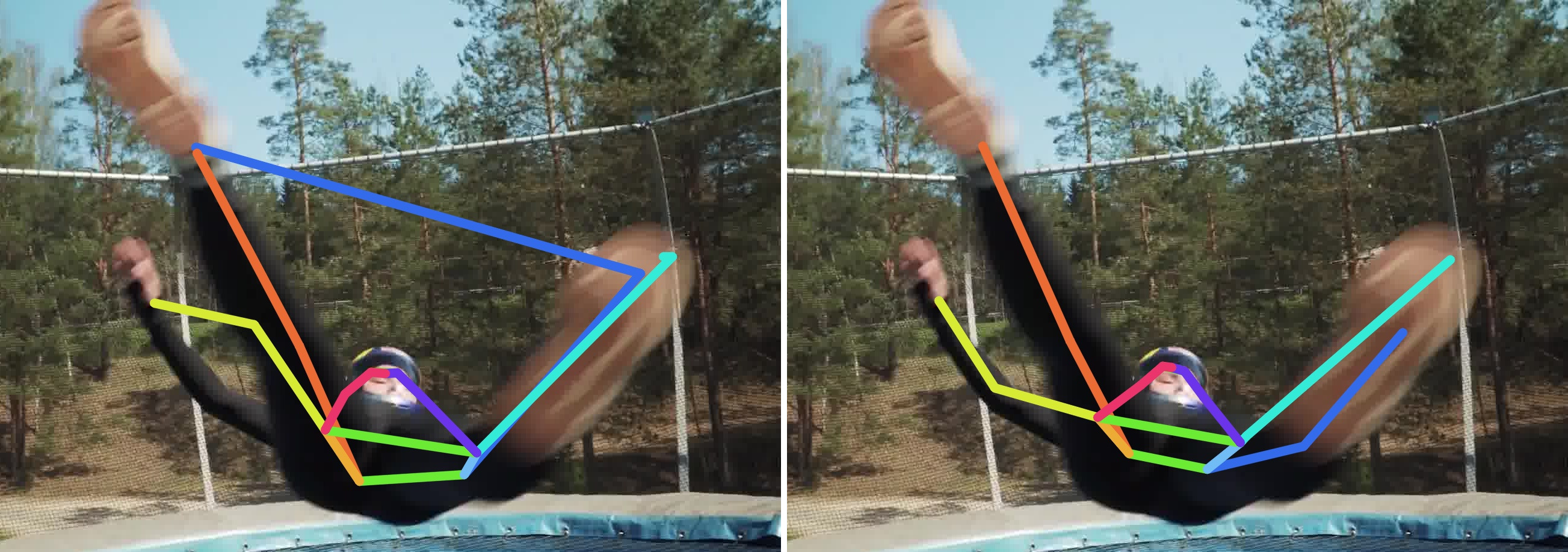}
  \end{subfigure}
  

  \begin{subfigure}{\linewidth}
    \centering
    \includegraphics[width=\linewidth]{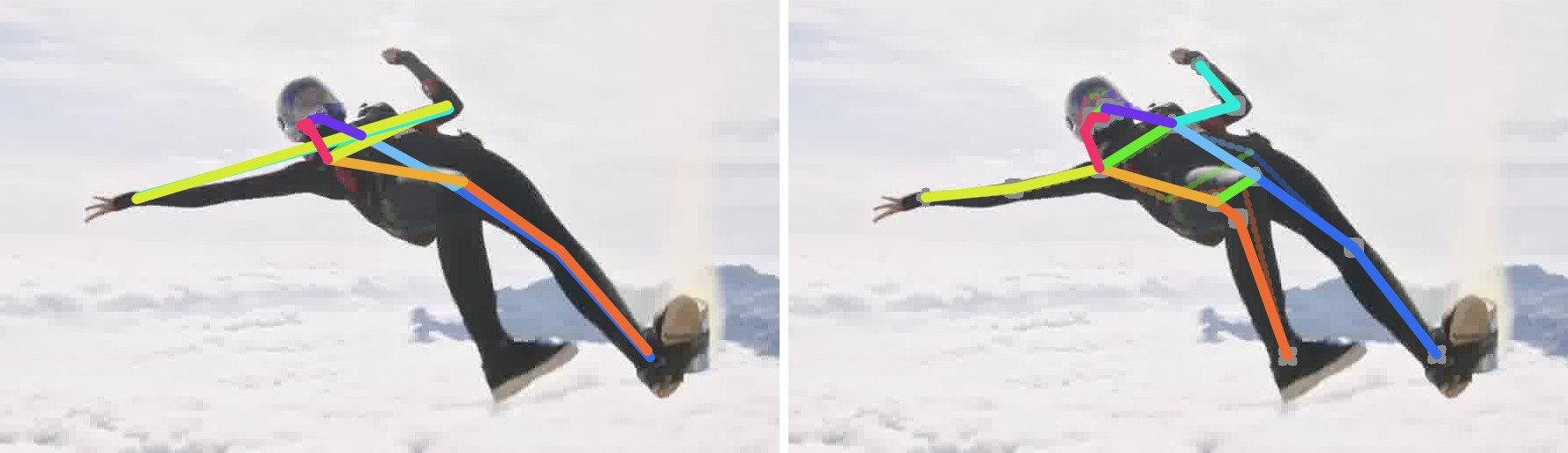}
  \end{subfigure}

  \begin{subfigure}{\linewidth}
    \centering
    \includegraphics[width=\linewidth]{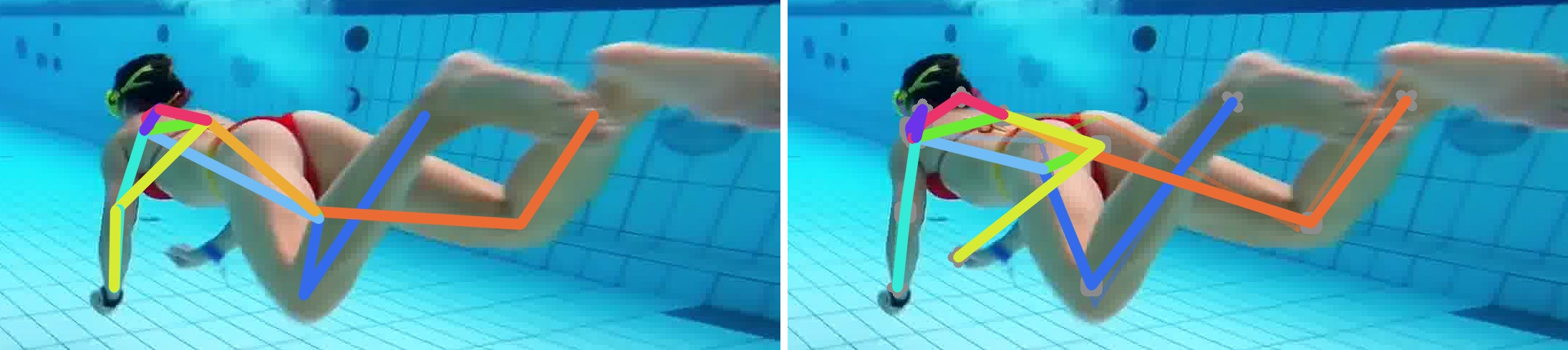}
  \end{subfigure}
  
  \caption{
  Examples from the RePo test set. ViTPose-s estimates when trained on COCO (left) and on RePoGen data (right).
  Colors as in \cref{fig:intro} -- 
  \colorbox{yellow}{right hand}, \colorbox{orange}{right leg}, \colorbox{SkyBlue}{left hand} and  \colorbox{blue}{\textcolor{white}{left leg}}
  }
  \label{fig:examples}
  
\end{figure}
\section{Method}
\label{sec:method}

\begin{figure*}[ht]
  \centering
  \begin{subfigure}{\linewidth}
    \centering
    \includegraphics[width=0.95\linewidth]{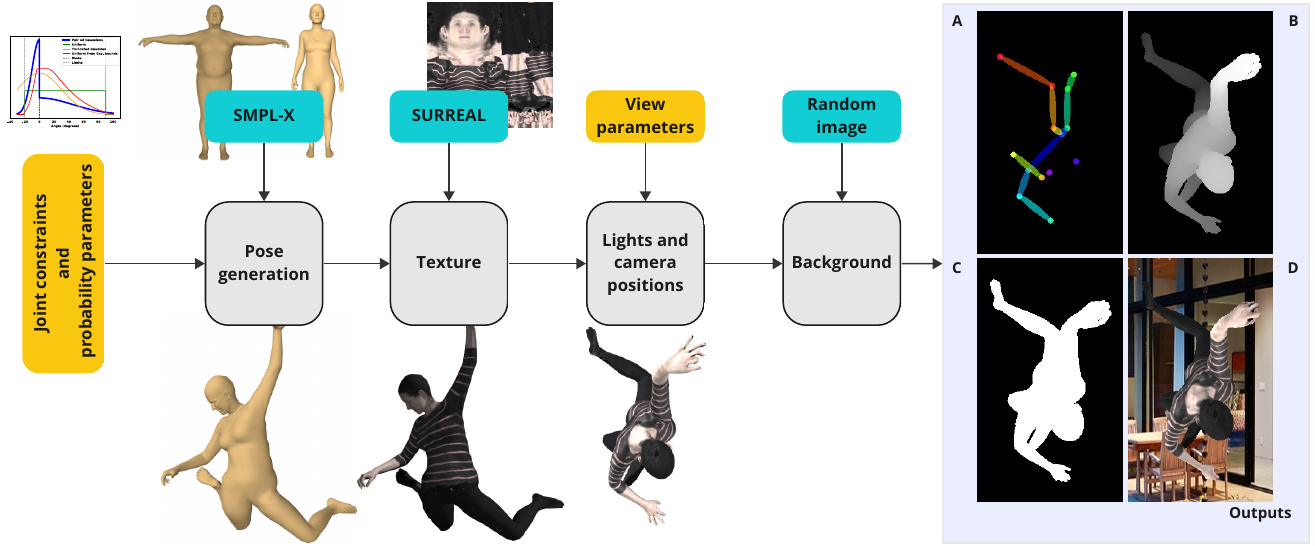}
  \end{subfigure}
  \caption{RePoGen synthetic data generation pipeline. All steps are detailed in \cref{sec:method}. The ground truth outputs of the method are (A) 2D and 3D keypoints, (B) the depth map, (C) the mask, and (D) an RGB image.}
  \label{fig:pipeline}
\end{figure*}

The proposed pipeline is outlined in the \cref{fig:pipeline}. The following paragraphs present a step-by-step walkthrough of the RePoGen data generation process, highlighting the main techniques employed to achieve pose control and generate diverse synthetic data.

Our method sample poses differently than prior works using 3D scans. RePoGen can generate completely new poses unseen in existing datasets or methods.
Although the anatomic plausibility of the generated poses is not guaranteed, we demonstrate that it is not a prerequisite for effective performance.

\subsection{Pose Generation}
\label{subsec:pose-gneration}



\begin{figure}[tb]
  \centering
  \includegraphics[width=\linewidth]{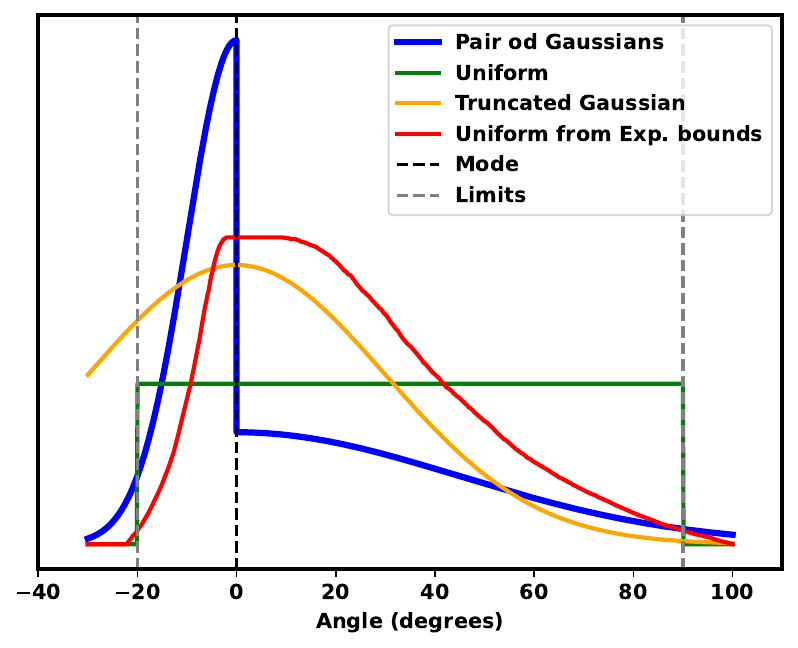}
    \caption{
    Set of tested joint rotation distributions used for pose generation.
    The pair of Gaussians is used in the final pipeline. Shown distribution is for left shoulder external and internal rotation.
    }
    \label{fig:distributions}
\end{figure}

Prior works like AMASS \cite{AMASS} or AGORA \cite{AGORA} focused on anatomically plausible synthetic poses. To guarantee that, they measured 3D body meshes using sophisticated techniques. The approach is limited to laboratory settings and cannot capture dynamic poses typical for sports where rare poses and views are the most prevalent. To overcome this limitation, we propose to sample poses from the space of \textit{bounded poses} $P^{bounded}$ (defined in \cref{def:p-bounded}) where we bound the rotation of each joint by its range of movement independently. The bounded space allows us to sample previously unseen poses in exchange for not guaranteed anatomical plausibility due to the joints' independence. The more thorough discussion is in the supplementary material.

\begin{definition}[$Space \ P^{bounded}$]
Space of all poses possible with a given kinematic chain defined by the SMPL model and joints ranges between ($\alpha_{j}$, $\beta_{j}$) for each joint \textit{j}. Angles of joints are independent.
\label{def:p-bounded}
\end{definition}

RePoGen leverages the SMPL-X \cite{SMPL-X} model with its 21 body joints to sample the $P^{bounded}$ space. In addition to the basic SMPL model \cite{SMPL}, SMPL-X also includes joints for hands and face. The rotation angles for the face and hand joints are randomly determined, as they do not influence the 17 COCO keypoints but increase the data variability.

By applying constraints on joints' rotation, a substantial portion of the pose space, primarily composed of unrealistic poses, is effectively eliminated. The remaining poses are highly likely to exhibit realistic characteristics, although some instances of mesh intersection may occur. However, these small-scale mesh intersections do not pose significant issues during training, as they effectively simulate minor body deformations within the rendered images.
The major advantage of our approach is the ability to generate rare poses that are not present in previous datasets.

\subsubsection*{Bounds and pose variance}

The space $P^{bounded}$ is defined by bounds ($\alpha_{j}$, $\beta_{j}$) for each joint. We chose the initial bounds according to the anatomical ranges of motion \cite{Kapandji1} used in physiotherapy.

When we multiply initial bounds by a constant, we further restrict the movement of the joint (and the space $P^{bounded}$) or allow for movement outside of the anatomical bounds, depending on the value of the constant. We call this constant \textit{pose variance}. The higher the pose variance, the more unusual poses RePoGen generates. Unusual poses add more information to the training process and raise the probability of sampling unrealistic poses. Lower pose variance creates poses more similar to the default one. Restricting the space $P^{bounded}$ reduces the risk of unrealistic poses and harms performance. Our experiments in \cref{subsec:ablation} show that some unrealistic poses do not hinder training as long as only a few of them exist.

\subsubsection*{Distribution}

To sample from $P^{bounded}$, we need a distribution function for each body angle. The distribution should meet the following conditions for $P^{bounded}$ to be as anatomically realistic as possible:

\begin{enumerate}
    \item The distribution is continuous and smooth, with exception of 0 (the default pose)
    \item The mode of the distribution is at 0 (to preserve the default pose)
    \item Bounds $\alpha_{j}$ and $\beta_{j}$ could be asymmetrical
    \item Distribution has low value in bounds (to avoid too many extreme poses)
\end{enumerate}

\cref{fig:distributions} shows various distributions that we tested. Each one is described in detail below. 

\textbf{Pair of Gaussians} (blue) is a combination of two normal distributions with discontinuity in the middle. The joint has the same probability of flexing or extending, and both bounds have a low probability.

\textbf{Uniform} distribution (green) does not have one mode to preserve the default pose and over-sample rotations around bounds, creating more extreme poses than pair of Gaussians.

\textbf{Truncated Gaussian} (orange) is smooth with one mode but is not asymmetrical, so one bound is favored over the other. 


\textbf{Uniform with exponentially distributed bounds} (red) combines exponential distribution with uniform. We first sample a pose variance from an exponential distribution and then the joint's rotation from a uniform distribution given by sampled pose variance. Although the mode is still not a single value, it performs better than plain uniform distribution because of the low probability of bounds. 

None of these fully meets all four conditions, but experiments show that the pair of Gaussians produces the best results on real-world images (see \cref{subsec:ablation}). This distribution allows us to generate pose angles centered around a standard pose, with unique and asymmetric ranges for each joint, while preserving the default pose by keeping the mode at zero.

Note that the same distribution function is shared by all joints, with each joint having different bounds.


\subsubsection*{Systematic search}

Searching the space $P^{bounded}$ for useful and realistic poses is not trivial, as deciding pose realisticity is challenging.
Automatic detection of pose realisticity through mesh intersections is complicated as the mesh self-intersects even when bending the joint. 
Mining for hard negatives leads to performance drop as we over-represent unrealistic poses.
Finding poses as a combination of previously seen ones, as in \cite{RarePoses}, does not explore new poses but merely interpolates between known ones.

The systematic search of the pose space remains an open problem.

\subsection{Texture}

Once the random pose is generated, we apply a randomized texture to the mesh. We utilize textures provided by the SURREAL project \cite{SURREAL} and do not differentiate between male and female textures. If no texture is applied (as examined in the ablation study in \cref{subsec:ablation}), we color the mesh to resemble natural skin tones. Therefore, the generated synthetic data exhibits variation in texture, contributing to a more realistic appearance.





\subsection{Random Background}

The final component for generating visually appealing images is the background. We incorporate a random image as the background and crop the rendered scene to a 1.25 multiple of the bounding box size. When selecting background images, we ensure that they depict environments where people are commonly observed. However, we refrain from including discernible individuals in the background, which could confuse the model since we do not focus on crowded scenes.

\subsection{Ground Truth Extraction}

The output of the pipeline includes not only the rendered RGB image but also the corresponding ground truth information. We first extract the depth map from the triangular mesh representation to obtain the ground truth. The depthmap is then used to generate a segmentation mask through thresholding. The segmentation mask defines the bounding box.

However, determining the visibility of joints is a complex process, as the joints of the SMPL-X model are positioned within the triangular mesh and are, therefore, always hidden from view in the rendered image. To address this, we define a neighborhood around each joint and consider the joint visible if at least one vertex from its respective neighborhood is visible in the image. The neighborhood size is proportional to the joint size and is determined based on the human annotation error defined in the OKS (object keypoint similarity) metric from the COCO dataset.
\section{Experiments}
\label{sec:experiments}

\subsection{Implementation Details}

To optimize computation power and time efficiency, we primarily conduct experiments using the ViTPose-s model unless otherwise specified. The training parameters align with the ViTpose model, with a batch size of 128 and a base learning rate 5e-5. We follow the training paradigm from \cite{PoseFES} and fine-tune the model pretrained on the COCO dataset.

To focus on analyzing and improving the pose estimation model, we utilize ground truth bounding boxes to crop individuals from the images to mitigate errors from detectors, particularly in extreme views.

All synthetic images used in experiments are generated exclusively through RePoGen, with a preference for the top or bottom views. Synthetic data from orbital views are not generated as they provide no notable improvement.

During training, the model is not exposed to any real extreme view images that are not present in the original COCO dataset. Instead, all additional data used for training purposes are synthetically generated. The model used for comparison with other approaches used 3,000 images. The ablation study was done using 1,000 images. 

\textbf{Rotation}. During training, we incorporate extensive rotation data augmentation of COCO and synthetic images. In experiments labeled as \textit{w/o rotation}, we follow the standard rotation augmentation up to $40^{\circ}$, while in other cases, we apply a rotation up to $180^{\circ}$.

\begin{table}[tb]
  \centering
  \begin{tabular}{@{}lr@{}}
    \toprule
    Dataset name & \# of poses \\
    \midrule
    WEPDTOF-Pose & 6749 \\
    PoseFES 1 & 736 \\
    RePo Bottom \small{(Val / Test / Test-Seq)} & 31/94/62 \\
    RePo Top & 91 \\
    \bottomrule
  \end{tabular}
  \caption{Number of annotated poses in evaluation datasets.}
  \label{tab:datasets}
\end{table}

\subsection{Datasets}
\label{sec:experiments-data}

\begin{table*}[tb]
    \centering
    \begin{tabular}{@{}l|l|rrrr@{}l}
        \toprule
        Model & Training data & RePo Bottom & RePo Top & WEPDTOF-Pose & PoseFES 1 &  \\
        \midrule
        HRNet & COCO & --- & --- & --- & 75.7 &  \\
              & THEODORE+ & --- & --- & --- & 76.1 & \makebox[-0pt]{}{$^\dagger$}\\
              & RePoGen top \small{(30 epochs)} & --- & --- & --- & 77.9 &  \\
              & RePoGen top & --- & --- & --- & \textbf{79.5} &  \\
        \midrule
        ViTPose-s & COCO & 35.1 & 40.9 & 47.1 & 68.9 & \\
                  & COCO \small{+ rot.} & 47.5 & 44.0 & 56.9 & 74.4 & \\
                  & AMASS top & 49.0 & 50.6 & 58.2 & 73.9 & \\
                  & AMASS bottom & 53.5 & 44.1 & 57.4 & 74.6 & \\
                  & RePoGen top & 46.3 & \textbf{55.7} & 57.9 & 75.2 & \\
                  & RePoGen bottom & \textbf{61.8} & 41.2 & 57.2 & 74.8 & \\
                  & RePoGen 50\% top + 50\% bottom & 53.9 & 55.6 & \textbf{58.5} & \textbf{75.5} & \\
        \midrule
        ViTPose-h & COCO & 69.2 & 62.0 & 78.0 & 77.8 & \\
                  & AMASS top & 73.0 & 66.0 & \textbf{84.3} & 80.0 & \\
                  & AMASS bottom & 80.5 & 63.7 & 83.7 & 79.6 & \\
                  & RePoGen top & 73.3 & \textbf{69.4} & \textbf{84.3} & \textbf{80.3} & \\
                  & RePoGen bottom & \textbf{81.1} & 63.0 & 83.7 & 79.9 & \\
        \bottomrule
    \end{tabular}
    \caption{
    AP on the RePo, WEPDTOF-Pose \cite{WEPDTOF-Pose}, and PoseFES \cite{PoseFES} datasets.
    Row - datasets used for training; column - datasets used for evaluation.
    THEODORE+, RePoGen, and AMASS mean adding synthetic data and rotation augmentation to the COCO dataset. 
    RePoGen and AMASS are using 3,000 images.
    The result marked ($^\dagger$) taken from \cite{PoseFES}.
    }
  \label{tab:SOTA}
\end{table*}

As far as we know, only two datasets exist for top-view pose estimation.
Therefore, we created a new real-world dataset to evaluate pose estimation in dynamic environments. We conduct experiments on the following datasets:

\textbf{COCO.}\cite{COCO} The standard dataset commonly used for human pose estimation. It contains approximately 250,000 annotated poses from various everyday activities. However, the COCO dataset includes very few images captured from extreme views.

\textbf{WEPDTOF-Pose.}\cite{WEPDTOF-Pose} WEPDTOF-Pose dataset published recently. It contains manual pose annotation for selected individuals from the WEPDTOF dataset \cite{WEPDTOF} used for person detection. The dataset comes in separate images, which suppresses various errors. For a fair comparison with other COCO-like datasets, we reverse-engineered the bounding boxes and used the COCO format so one image contains multiple people. Although this is the biggest dataset for top-view pose estimation, the quality of images is low due to small bounding boxes. The dataset predominantly contains orbital view images due to the fisheye transformation.

\textbf{PoseFES.}\cite{PoseFES} PoseFES is a manually annotated dataset captured by a ceiling-mounted fisheye camera, serving as the solely available top-view dataset for human pose estimation. PoseFES consists of two sequences: one focusing on two well-separated individuals, while the second involves multiple people interacting and creating challenging scenarios with occlusions. We primarily utilize the first sequence for testing to align with our research focus on single-person human pose estimation. As with the WEPDTOF-Pose, the dataset predominantly contains orbital view images due to the fisheye transformation.

\textbf{RePo Bottom} Since no existing datasets specifically cater to bottom-view data, we created a new dataset called RePo (RarE POses) to evaluate our approach. The dataset consists of images extracted from various sports videos obtained from YouTube. The most common sports featured are swimming, climbing, and skydiving. The Val and Test sets possess similar structures derived from comparable videos, while the Test-Seq set comprises consecutive frames from one specific video of the pole vault. We employ the Test-Seq set to demonstrate that substantial rotations of the person often accompany extreme views. Examples of real images from the new dataset are in the \cref{fig:examples}.

\textbf{RePo Top} Similar to the Bottom datasets, the Top dataset is collected from sports videos focusing on the top-view perspective. It serves mainly as a validation set during the top-view training phase. The Top Val is also part of the new RePo dataset.

For further reference, a summary of the introduced datasets is presented in the \cref{tab:datasets}.

\textbf{Metrics.} All experiments were conducted following the COCO-style settings. The evaluation metric used was OKS-based AP (average precision), as specified in the COCO dataset \cite{COCO}.

\subsection{Comparison with baseline}

The comparison \cref{tab:SOTA} illustrates the performance comparison between the baseline model, fine-tuning on THEODORE+ \cite{PoseFES}, adding synthetic data using AMASS poses \cite{AMASS} and the proposed RePoGen method.

The first part of the table illustrates the performance gain compared to the THEODORE+.
We conducted fine-tuning of the HRNet \cite{HRNet} model from the MMPose \cite{MMPose} model zoo following the same procedure as described by Yu et al. \cite{PoseFES}.
We observed that surpassing the prescribed 30-epoch fine-tuning, as mentioned in \cite{PoseFES}, led to further improvements in performance.
RePoGen achieves superior results despite utilizing significantly fewer data, incorporating 3000 synthetic images compared to 160,000 THEODORE+ images.
Our method is only compared with the PoseFES dataset because we lack access to the model from \cite{PoseFES}.

The second part of the table compares our pose generation technique with AMASS poses acquired through 3D scans. To ensure a fair assessment, we sampled the same number of poses as in RePoGen — 3,000 for the results presented in \cref{tab:SOTA} and 1,000 for the ablation study.
RePoGen surpasses the performance of all previous methods, especially in the context of the dynamic RePo dataset, where it leverages less common poses than AMASS.
Our method demonstrates a marginal but consistent edge over the competition on standard datasets PoseFES and WEPDTOF-Pose.
However, including rotation augmentation to the COCO dataset substantially improves performance on PoseFES and WEPDTOF-Pose datasets, implying that these benchmarks predominantly evaluate fisheye performance rather than top-view scenarios.

Incorporating synthetic data from the bottom view enhances the model's performance on the bottom and top view, suggesting a similarity between the two extreme view domains. Similarly, training with synthetic data from the top-view demonstrates improvements across top-view and bottom-view scenarios.

The last part of \cref{tab:SOTA} shows that previous claims hold even for the biggest VitPose-h model.

\subsection{Ablation Study}
\label{subsec:ablation}

We analyze and evaluate the influence of each component individually, as described in the following paragraphs. The strong rotation augmentation is consistently applied throughout the ablation study, and unless otherwise specified, 1000 RePoGen are used for experimentation.

\begin{table}[tb]
  \centering
  \begin{tabular}{@{}lrr@{}}
    \toprule
    \# of images & Bottom Test & Bottom Test-Seq \\
    \midrule
    0    &  47.5 & 75.2 \\
    500  &  54.1 & 86.1 \\
    1000 &  59.1 & 89.0 \\
    3000 &  \textbf{61.8} & \textbf{90.5} \\
    5000 &  58.8 & 86.1 \\
    \bottomrule
  \end{tabular}
  \caption{
  AP on the RePo Bottom dataset; training with different numbers of RePoGen images.
  }
  \label{tab:number-of-images}
\end{table}

\begin{table}[tb]
  \centering
  \begin{tabular}{@{}lr@{}}
    \toprule
    Distribution & Bottom Test \\
    \midrule
    pair of Gaussians & 59.1  \\
    Truncated Gaussian & 55.6 \\
    Uniform \small{(fixed pose variance)} & \textbf{59.2} \\
    Uniform \small{(pose variance from $Exp(\lambda)$)} & 58.5 \\
    AMASS poses & 54.4 \\
    \bottomrule
  \end{tabular}
  \caption{
  AP on the RePo Bottom dataset; training with different joint distributions.
  }
  \label{tab:distributions}
\end{table}

\begin{table}[tb]
  \centering
  \begin{tabular}{@{}lcc@{}}
    \toprule
    RePoGen version & Bottom Test & Bottom Test-Seq \\
    \midrule
    baseline & 59.1 & \textbf{89.0}\\
    w/o rotation & 45.9 & 72.3 \\
    w/o background & 56.2 & 85.2 \\
    w/o texture & \textbf{59.5} & 88.2 \\
    \bottomrule
  \end{tabular}
  \caption{
  Ablation study.
  Training without various components - AP comparison on the Bottom dataset of RePo.
  }
  \label{tab:ablation}
\end{table}

\begin{figure}[tb]
  \centering
  \includegraphics[width=\linewidth]{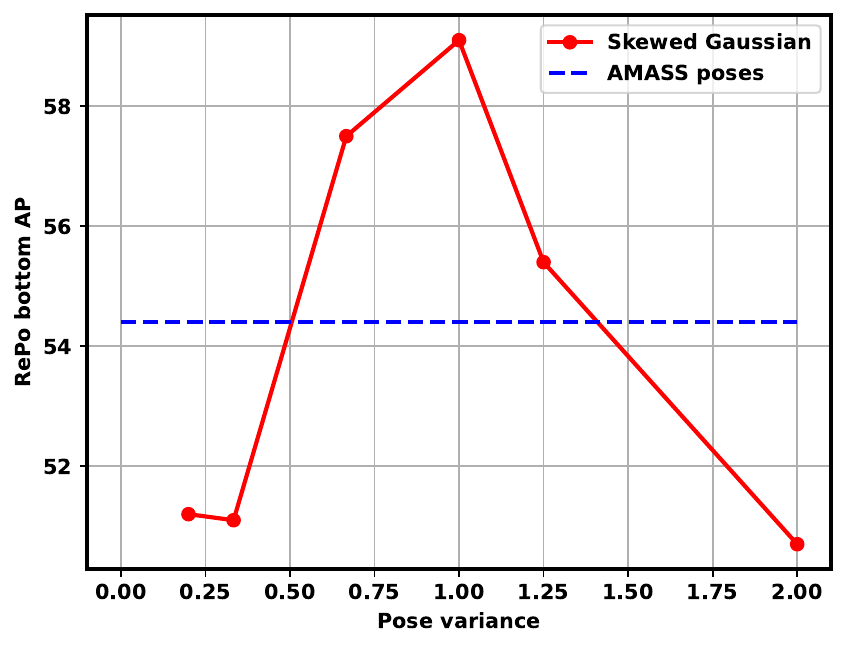}
    \caption{
    AP on the Bottom dataset of RePo; training with pair of Gaussians distribution with different values of pose variance. Low pose variance means that poses are not diverse enough, while high numbers signify too unrealistic poses.
    }
    \label{fig:pose_simplicity}
\end{figure}

\textbf{Number of images.} The \cref{tab:number-of-images} provides insights into the impact of adding additional images to the COCO dataset. With the COCO train set already containing over 200,000 poses, adding 5,000 images represents approximately 2\% of the dataset, resulting in minimal impact on training time. Remarkably, even including as few as 500 images yields noticeable improvements. Saturation is observed at around 3,000 images, beyond which further additions may have a marginal negative effect on performance, probably due to the domain gap and overfit to the synthetic data.


\textbf{Rotation.} Incorporating stronger rotation yields significant performance improvements. The effect is particularly pronounced in the Test-Seq set, where the presence of views adjacent to the extreme ones amplifies the difference even further. Even without rotation, our approach outperforms the off-the-shelf model, highlighting the importance of including extreme view data in the training. Consequently, it is advisable always to employ rotation data augmentation up to 180$^{\circ}$ for applications involving pose estimation in videos with extreme views. Experimental backup is in \cref{tab:SOTA} and \cref{tab:ablation}.

\textbf{Pose sampling and distributions.}
We present a comparative analysis of three basic techniques for pose generation: the baseline approach using a pair of Gaussians distribution, uniform distribution, and a truncated Gaussian distribution in \cref{tab:distributions}.
Additionally, we combine these basic distributions into combined ones described in \cref{sec:method} and add a comparison with AMASS \cite{AMASS} poses.
The experimental outcomes reveal that poses sourced from AMASS exhibit the least favorable performance, no matter whether using 500, 1,000, or 3,000 images.
The best performance is with plain uniform distribution, but contrasting results were observed when training on the top view and evaluating on the PoseFES dataset.
The pair of Gaussians performs almost the same and shows no such problems.

\textbf{Pose variance.}
The \cref{fig:pose_simplicity} inspects the performance when sampling from pair of Gaussians with various values of pose variance. As explained in \cref{sec:method}, high pose variance means a higher probability of sampling joint rotation on the edge of possibility, leading to more unrealistic poses. On the other hand, low pose variance does not allow for sampling uncommon poses, leading to low performance. Therefore, there is a tradeoff between realisticity and exploration of unseen poses. The \cref{fig:pose_simplicity} shows that having too many unrealistic poses is as bad as having too few. The blue dashed line illustrates the performance of AMASS poses.

\textbf{Pipeline design.}
\cref{tab:ablation} corroborates the efficacy of various design decisions in the pose generation methodology.
In particular, incorporating background images leads to a slight improvement in performance, whereas the introduction of random textures does not produce notable enhancements.
This implies that data realism might not be essential in this scenario.
However, excluding texture negatively affects performance in different datasets, thus its inclusion was maintained.
The table further highlights the significance of rotation data augmentation.


\section{Conclusions}
\label{sec:conclusions}

We presented a novel method for generating synthetic images (RePoGen) with accurate human pose ground truth by incorporating constraints on joint rotation. We trained a state-of-the-art model on the COCO dataset enhanced by RePoGen data to improve performance in extreme views.
The key findings can be summarized as follows:

\begin{enumerate}


\item Including a small number of synthetic training samples with extreme views significantly improved extreme view pose estimation.

\item Sampling from the bigger $P^{bounded}$ space seems preferable to 3D scans as it allows exploration of new poses even while risking sampling unrealistic ones.

\item Stronger rotation data augmentation proved crucial, particularly for views adjacent to extreme viewpoints. This augmentation technique is recommended especially for fisheye ceiling-mounted cameras.

\item The pose estimation performance increased when synthetic data closely resembled the poses observed in the target domain.

\end{enumerate}

The next step would be utilizing the proposed model to pre-annotate a larger dataset of extreme views from sports using a human-in-the-loop approach to enable further investigation into the challenges arising from extreme poses. By delving deeper into these complexities, future research endeavors can enhance the understanding and performance of pose estimation in extreme-view scenarios. Furthermore, the annotated dataset comprising almost 200 images of the bottom view and nearly 100 images of the top view, primarily sourced from sports activities, will be made publicly available, contributing to the advancement of the field. Another improvement would be to search the pose space systematically for effective hard-negative mining.  

RePoGen data improves human pose estimation even with limited realism. We experimented with ControlNet \cite{ControlNet} to improve the data realism. ControlNet only produces realistic results for orbital views and fails catastrophically for extreme or rare poses. We believe RePoGen data could be suitable for ControlNet fine-tuning. 


\textbf{Acknowledgements.} This work was supported by the Technology Agency of the Czech Republic project No. SS05010008, Ministry of the Interior of the Czech Republic project No. VJ02010041 and Czech Technical University student grant SGS23/173/OHK3/3T/13.
{
    \small
    \bibliographystyle{ieeenat_fullname}
    \bibliography{main}
}

\maketitlesupplementary
\setcounter{page}{1}    

\section{Pose spaces analysis}
\label{sec:poses}

\subsection{Definitions}
\label{subsec:Poses-definitions}

The spaces associated with pose estimation can be conceptualized through the popular SMPL model \cite{SMPL}. The fixed kinematic chain allows us to describe each pose by the rotations of its joints. Every pose inherent to the SMPL model can be uniquely denoted as a vector positioned in a 63-dimensional space (21 joints, 3 axes of rotations each), each dimension bounded in the range of ($-\pi$, $\pi$). We call this space $P^{all}$. 

\begin{definition}[$P^{all}$]
Space of all poses possible with a given kinematic chain defined by the SMPL model and joints ranges between ($-\pi$, $\pi$).
\label{def:p_all}
\end{definition}

Once the $P^{all}$ is defined, we can concretize other spaces to analyze prior synthetic data generation techniques.

\begin{definition}[$P^{bounded}$]
Space of all poses possible with a given kinematic chain defined by the SMPL model with joints ranges between ($\alpha_{j}$, $\beta_{j}$) for each joint \textit{j}. Angles of joints are independent.
\end{definition}

\begin{definition}[$P^{anatomical}$]
Space of all poses that at least one human can achieve.
\end{definition}

\begin{definition}[$P^{AMASS}$]
Space of all poses captured in the AMASS dataset.
\end{definition}

\begin{figure*}[bt]
  \centering
  \begin{subfigure}{\linewidth}
    \centering
    \includegraphics[width=0.95\linewidth]{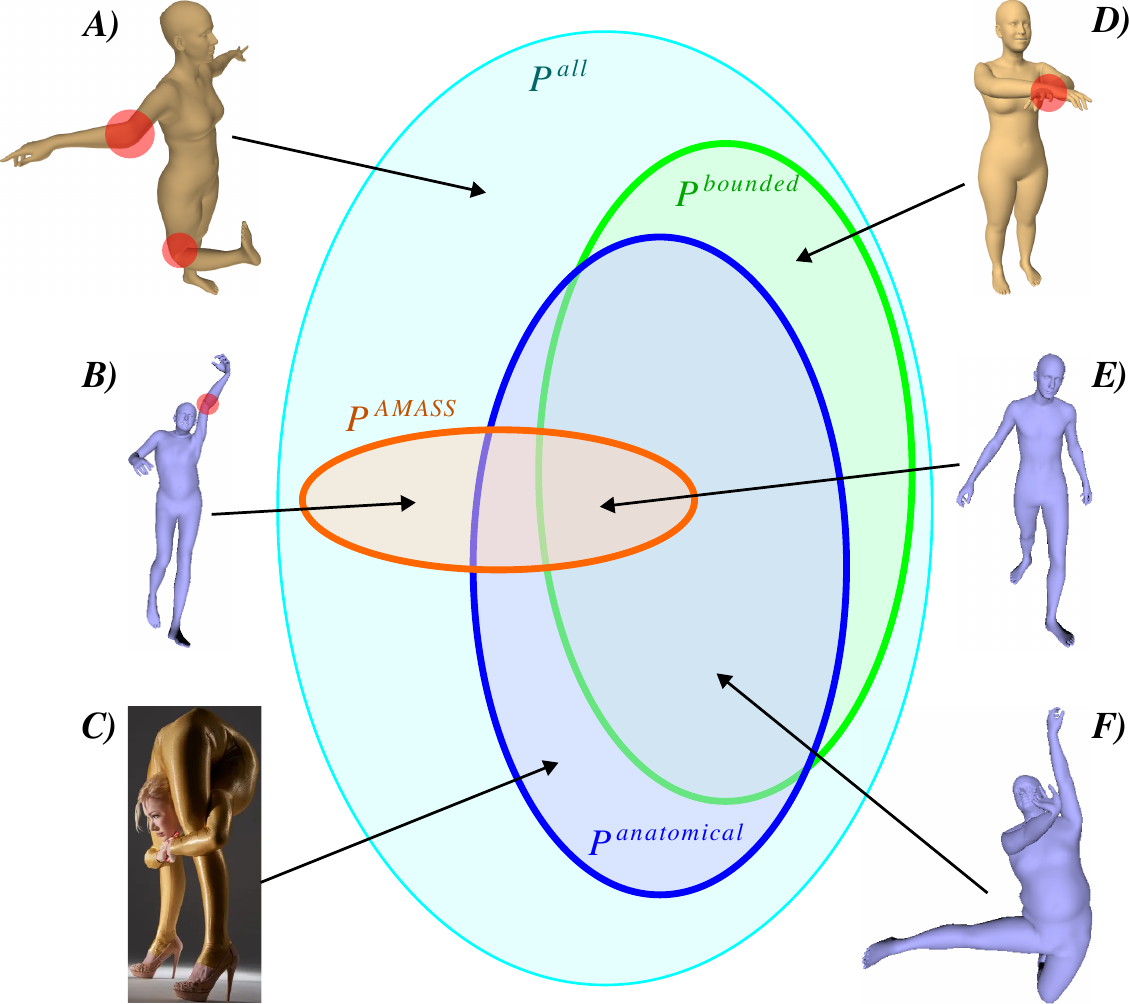}
  \end{subfigure}
  \caption{Visualization of described pose spaces and examples of poses from them. Pose spaces are described in detail in \cref{subsec:Poses-definitions}. The examples are A) impossible pose, B) pose from AMASS dataset with impossible joint rotation, C) anatomical but not bounded pose, D) bounded pose with self-intersection, E) AMASS pose with low pose variance, F) RePoGen pose not present in the AMASS dataset. Impossible rotations and self-intersections emphasized by red circles.}
  \label{fig:pose-spaces}
\end{figure*}

All defined spaces along with examples are visualized in the \cref{fig:pose-spaces}.

In the space $P^{all}$, we use Euclidean distance to measure the similarity of the two poses. Using the Euclidean distance is not trivial; we justify it in two steps.

(1) With the given kinematic chain of the SMPL model, the two poses differ only by angles of joints. In the rendered image, a camera viewpoint is crucial for human-perceived pose similarity (and also 2D pose similarity). To discard the influence of the camera, we treat each element of the pose vector independently with the same weight.

(2) Euclidean distance would not work for angles on the border between $-\pi$ and $\pi$ where Euclidean distance is biggest while the angle distance is low. We theorize that most human joints have a range of movement smaller than $\pi$, and the mentioned overflow of the distance will not happen between poses close to the human anatomy. It is worth noting that the Euclidean distance would fail for two unbounded poses.

Using Euclidean distance allows us to use off-the-shelf algorithms like k-means clustering or dimensionality reduction techniques for further analysis.  

\subsection{Synthetic data generation through the pose spaces}

Previous works like AMASS \cite{AMASS} focus on generating anatomically plausible data from the space $P^{anatomical}$. They do so by fitting the SMPL model into 3D scans of humans. The approach has two challenges. 

First, an error can occur during the SMPL model fitting, and the resulting pose would be outside of the $P^{anatomical}$. An example of such an error is the \cref{fig:pose-spaces} B) where the left elbow is rotated by more than 45$^{\circ}$ along the axis where no rotation is anatomically possible.

The second and bigger challenge is that the approach requires many expensive 3D scans of people in various poses. Due to the nature of the 3D scanning techniques, most poses are not dynamic (jumping, running, etc.), and the researcher in the lab must design rare poses. There is a high probability that a lot of the $P^{anatomical}$ space remains unsampled.

On the other hand, sampling poses from the $P^{bounded}$ space has its challenges. Foremost, how to design the bounds and sample the space to balance rare poses (as is the case of \cref{fig:pose-spaces} F) with real-world ones. The bounds should not be too tight to allow sampling of very rare poses but also not too loose to generate a lot of anatomically impossible poses (like the pose D) in \cref{fig:pose-spaces}). The optimal bounds would cause spaces $P^{anatomical}$ and $P^{bounded}$ to overlap almost completely. There will always be poses impossible to capture by the $P^{bounded}$ - for example, contortionist visualized in \cref{fig:pose-spaces} C).

\subsection{RePoGen vs. AMASS}

When comparing the RePoGen-generated data with the AMASS dataset through the spaces introduced in \cref{subsec:Poses-definitions}, we can measure their similarity (using the Euclidean distance) and the number of rare poses.

\begin{figure}[tb]
    \centering
    RePoGen \hfill Transitions \\
    \begin{subfigure}{0.18\linewidth}
        \includegraphics[width=\linewidth]{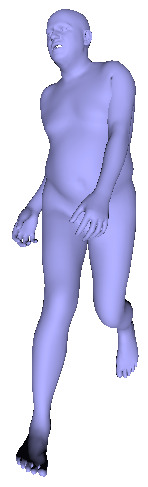}
    \end{subfigure}
    \hspace{3cm}
    \begin{subfigure}{0.225\linewidth}
        \includegraphics[width=\linewidth]{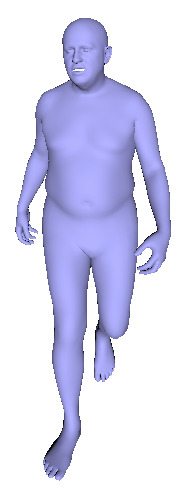}
    \end{subfigure}

    RePoGen \hfill Human4D \\
    \begin{subfigure}{0.43\linewidth}
        \includegraphics[width=\linewidth]{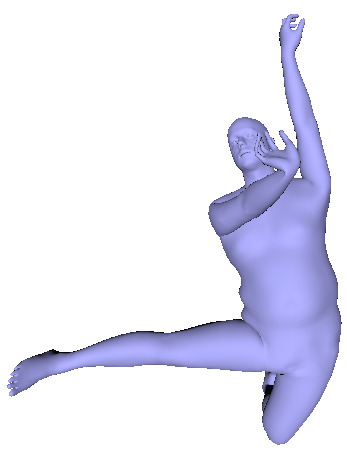}
    \end{subfigure}
    \hfill
    \begin{subfigure}{0.25\linewidth}
        \includegraphics[width=\linewidth]{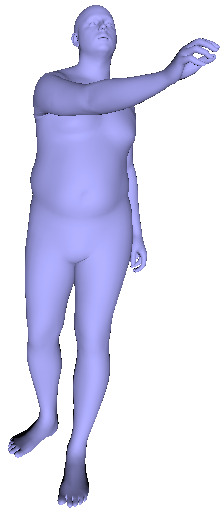}
    \end{subfigure}
    \hspace{0.5cm}
    
    \caption{Selected RePoGen poses and their most similar AMASS counterpart. The first row shows two similar poses generated by the RePoGen and AMASS pipelines, while the second row proves the RePoGen pipeline generates poses unseen in AMASS. Each AMASS pose is labeled by its respective subset.}
    \label{fig:min-and-max-poses}
\end{figure}

To measure the ratio of rare poses as in the \cite{RarePoses}, we clustered AMASS poses using the k-means. Poses further than the threshold from all cluster centers are classified as rare. Setting up the threshold such that the AMASS dataset has 5\% of rare poses results in over 90\% of rare poses in the RePoGen data. RePoGen data are, therefore, different from AMASS data by a big margin. RePoGen data are rare and usually completely new instead of weighting rare poses as in \cite{RarePoses}. To justify using the RePoGen data instead of AMASS, the \cref{fig:min-and-max-poses} shows the most similar AMASS pose to selected RePoGen poses. We can see that while RePoGen generates poses similar to standard ones, it also generates ones not in the AMASS dataset.

\section{RePo Dataset}
\label{sec:RePo}

Here we describe a new RePo dataset of manually annotated real images. The dataset focuses on extreme poses in top and bottom views typically encountered in sports. Images come primarily from public sports videos on YouTube. The dataset is split into two parts - one focusing on the bottom view with 187 images, the other focusing on the top view with 91 images. Each part is divided into sets described in Tab. 1 in the paper.

One professional annotator annotated the whole dataset. We created a custom annotation environment allowing for an easier understanding of the scene necessary for annotating extreme views. Since the visibility of the joints is not defined in detail in the COCO dataset \cite{COCO}, we defined it as follows:

\textbf{Visibility 0.} The keypoint is not visible, and we cannot reliably tell its precise location. The words \textit{reliably} and \textit{precise} are crucial in situations where a keypoint is not visible. We can guess its location from the context but cannot be sure the guess was correct.

\textbf{Visibility 1.} The keypoint is not visible, but we can reliably tell its precise location from the context and other keypoints.

\textbf{Visibility 2.} The keypoint is visible in the image.

Further, as the extreme views pose additional challenges, we stick to the annotation of joint projection to the image plane if it is unambiguous. A typical example would be the ankle which is rarely visible from the bottom view (we see the heel instead). Without this relaxation of definition, almost no keypoints would be annotated as visible (visibility 2).

Examples of images from all sets of the dataset are in the \cref{fig:RePo-bottom-test}, \cref{fig:RePo-bottom-val} and \cref{fig:RePo-bottom-seq}. 

\section{RePoGen Dataset}
\label{sec:RePoGen}

\begin{figure}[tb]
  \centering
  \begin{subfigure}{0.3\columnwidth}
      \includegraphics[width=\textwidth]{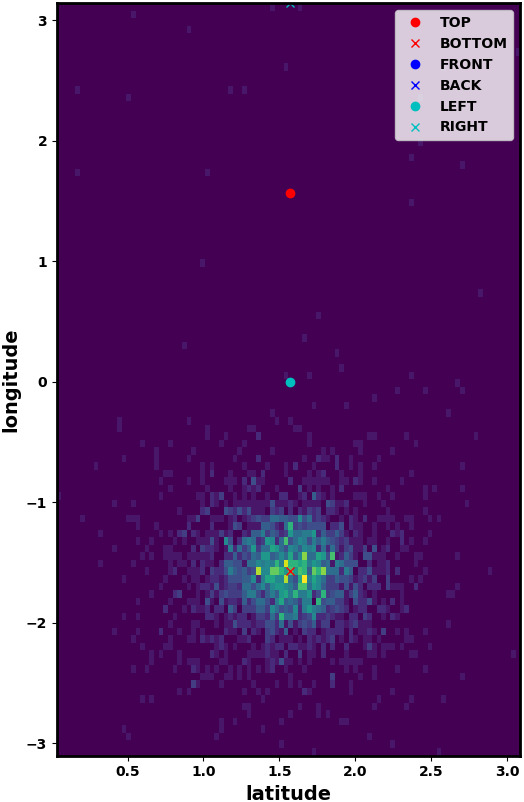}
      \caption{\small{(bottom)}}
  \end{subfigure}
  \hfill
  \begin{subfigure}{0.3\columnwidth}
      \includegraphics[width=\textwidth]{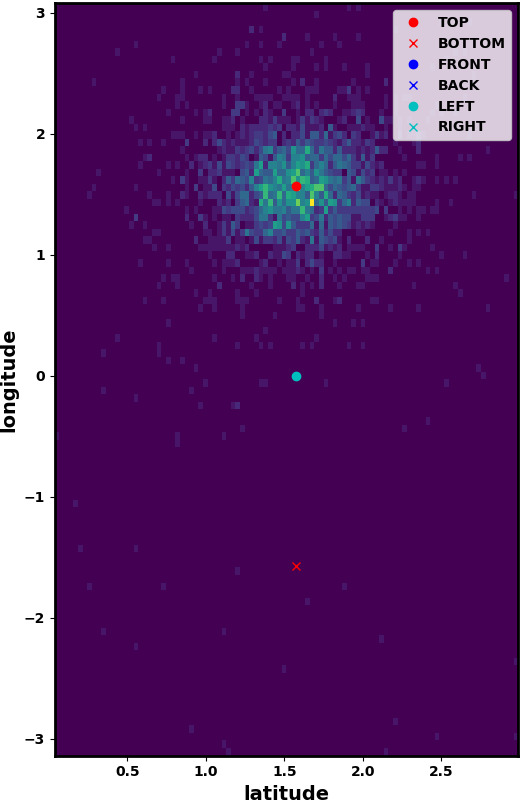}
      \caption{\small{(top)}}
  \end{subfigure}
  \hfill
  \begin{subfigure}{0.3\columnwidth}
      \includegraphics[width=\textwidth]{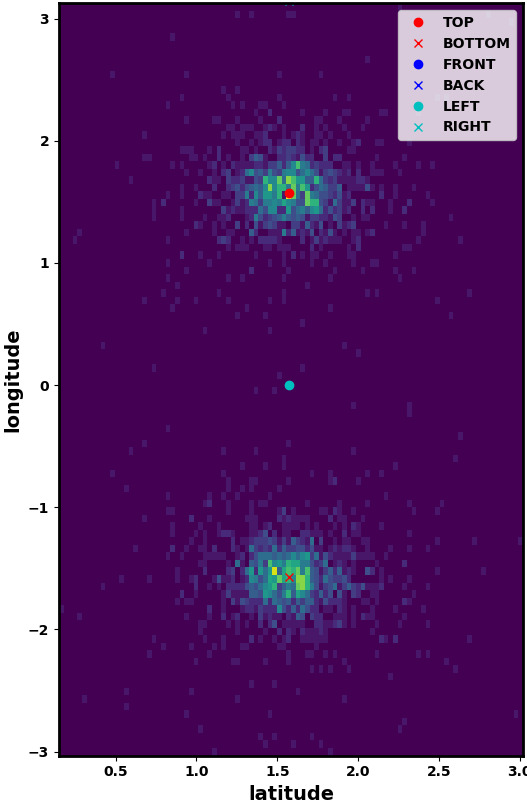}
      \caption{\small{(top+bottom)}}
  \end{subfigure}
   \caption{
   Camera viewpoint distributions for various RePoGen datasets. Vertical axis - latitude, horizontal - longitude.
   }
   \label{fig:RePoGen-distributions}
\end{figure}

The RePoGen dataset was created with the proposed RePoGen method and was used to train the best-performing model. There are 3 variants of the RePoGen dataset, all meeting the description in \cref{tab:params}. The parameter distinguishing the 3 variants is the camera viewpoint distribution. The RePoGen \small{(bottom)} and RePoGen \small{(top)} are sampled with a normal distribution centered around the bottom view and top view respectively. The RePoGen \small{(top+bottom)} is sampled from a combination of these two distributions. Visualization is in the \cref{fig:RePoGen-distributions}, where 3D coordinates are projected to the latitude and longitude.

\begin{figure}[tb]
  \centering
   \includegraphics[width=\linewidth]{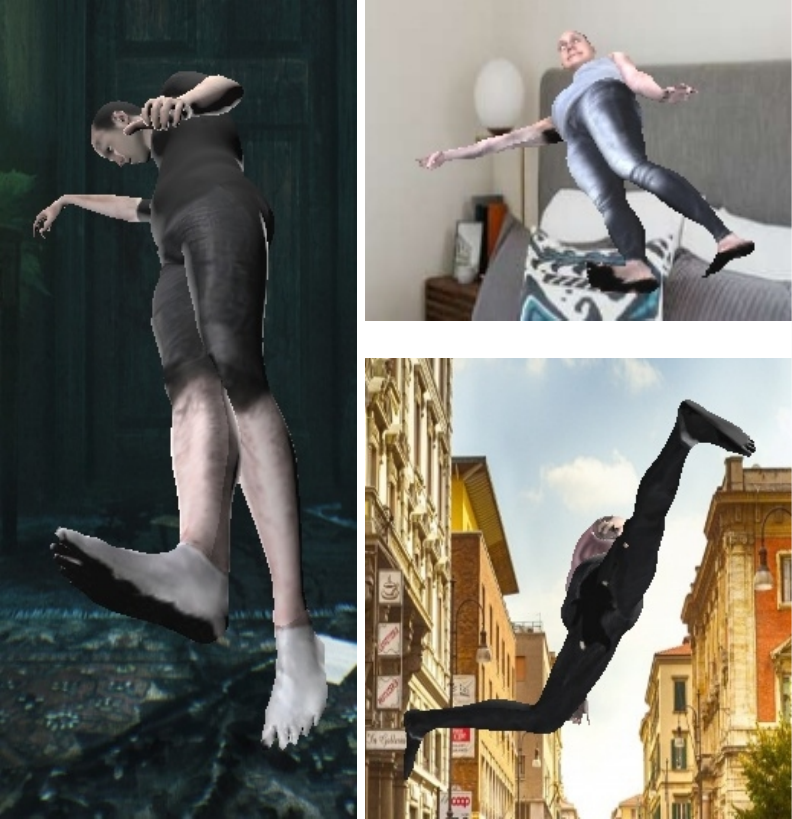}
   \caption{
   Images from the RePoGen \small{(bottom)} dataset.
   }
   \label{fig:RePoGen-bottom}
\end{figure}

\begin{figure}[tb]
  \centering
   \includegraphics[width=\linewidth]{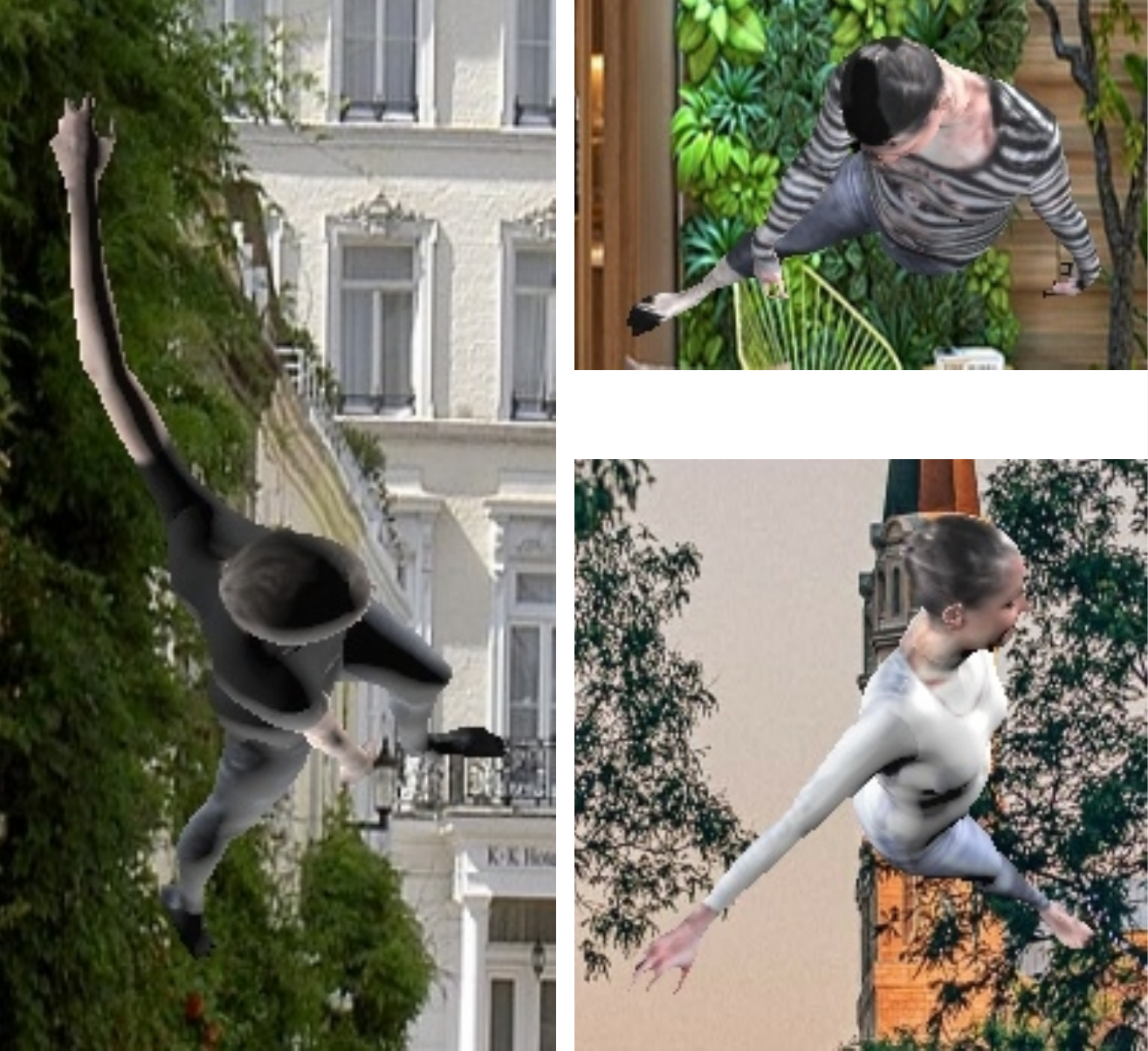}
   \caption{
   Images from the RePoGen \small{(top)} dataset.
   }
   \label{fig:RePoGen-top}
\end{figure}

The \cref{fig:RePoGen-bottom,fig:RePoGen-top} contain images from the RePoGen dataset.

\begin{table}[tb]
  \centering
  \begin{tabular}{@{}lr@{}}
    \toprule
    parameter & value \\
    \midrule
    number of poses & 1 500 \\
    number of views per pose & 2 \\
    pose variance & 1.0 \\
    with texture & \cmark \\
    with background & \cmark \\
    default SMP pose & \xmark \\
    distribution & skewed Gaussian \\
    \bottomrule
  \end{tabular}
  \caption{Parameters used for RePoGen dataset generation.}
  \label{tab:params}
\end{table}

\section{Additional results}
\label{sec:results}

\begin{figure}[tb]
  \centering
  \begin{subfigure}{\linewidth}
    \centering
    \includegraphics[width=\linewidth]{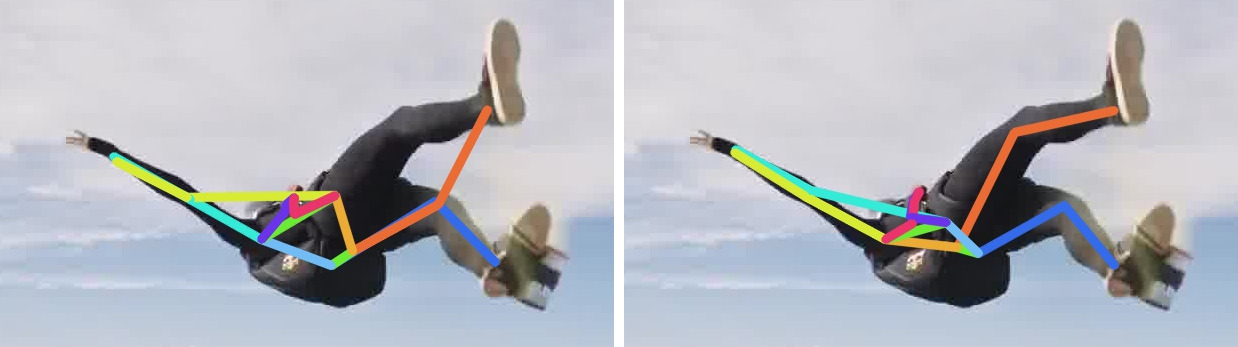}
  \end{subfigure}
  
    
  
  \begin{subfigure}{\linewidth}
    \centering
    \includegraphics[width=\linewidth]{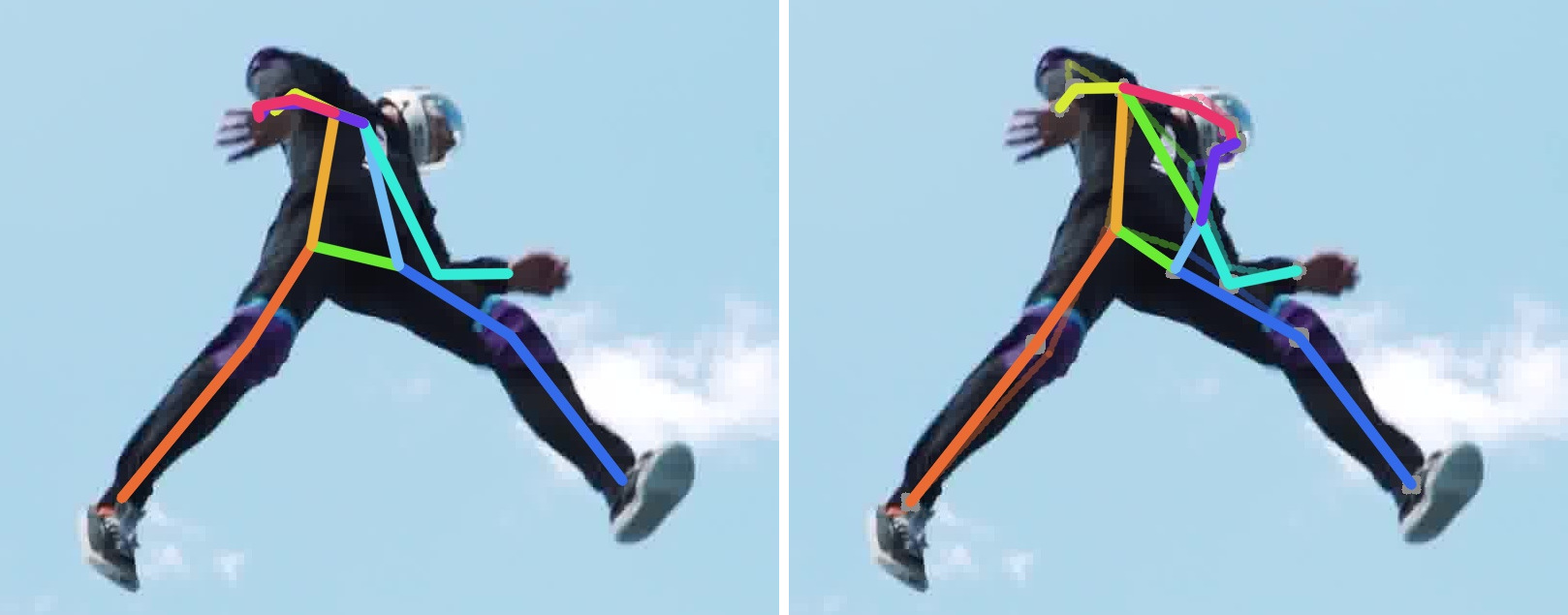}
  \end{subfigure}
  
  \begin{subfigure}{\linewidth}
    \centering
    \includegraphics[width=\linewidth]{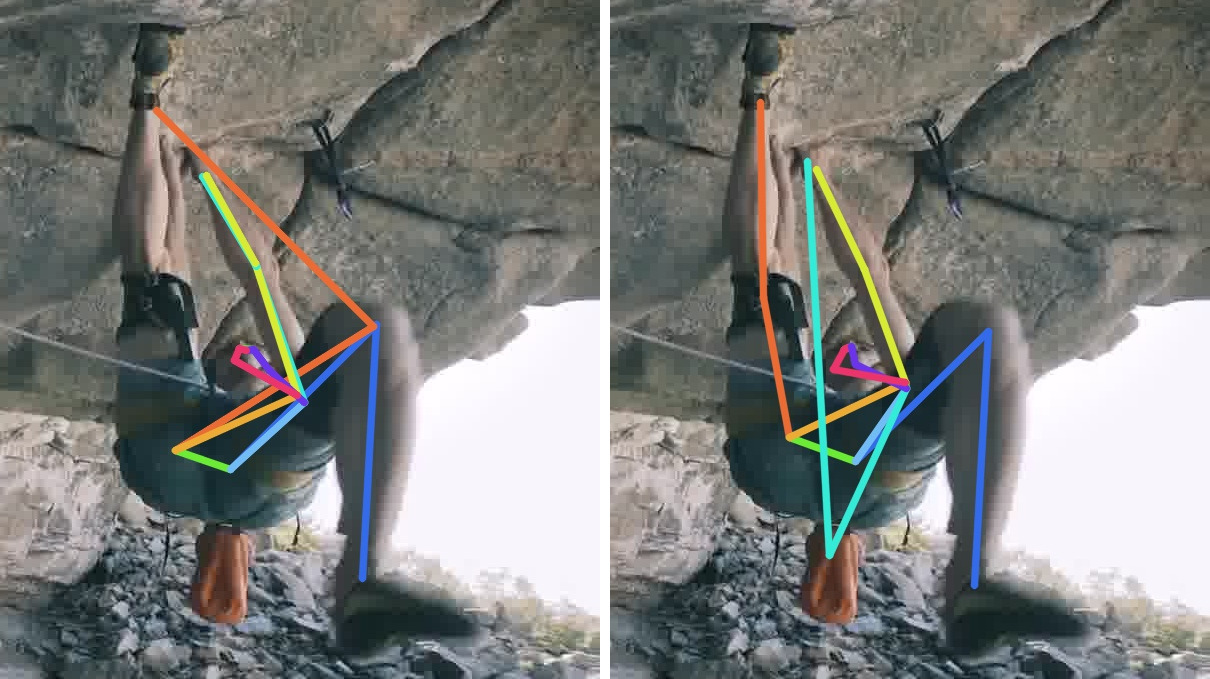}
  \end{subfigure}
  
  \begin{subfigure}{\linewidth}
    \centering
    \includegraphics[width=\linewidth]{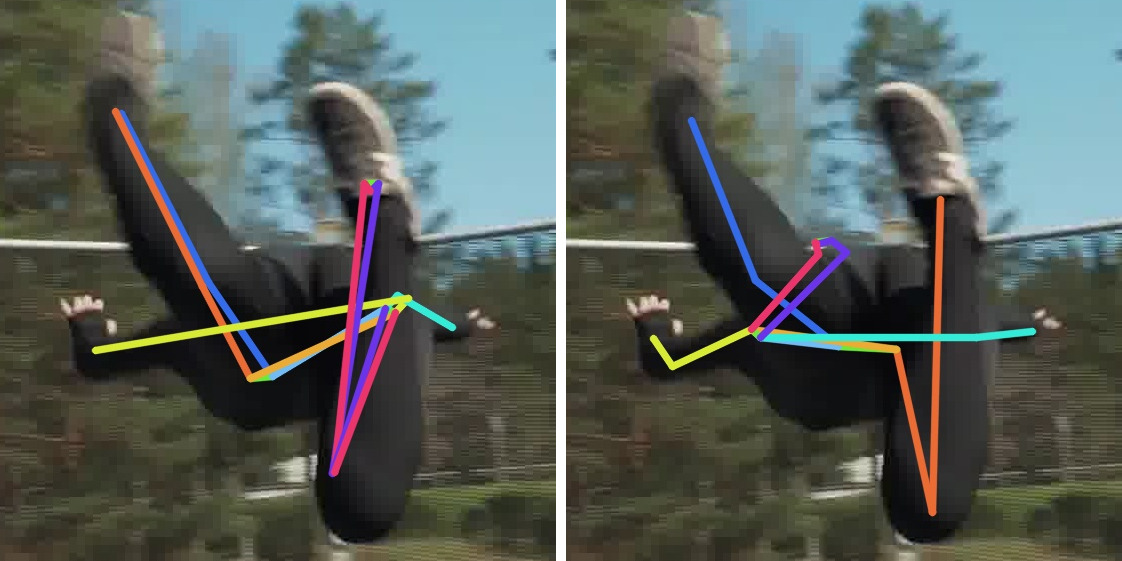}
  \end{subfigure}
  
  \begin{subfigure}{\linewidth}
    \centering
    \includegraphics[width=\linewidth]{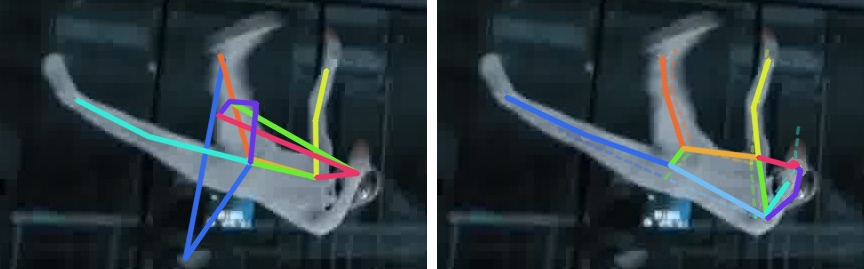}
  \end{subfigure}
  
  \begin{subfigure}{\linewidth}
    \centering
    \includegraphics[width=\linewidth]{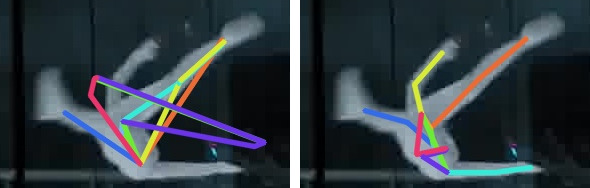}
  \end{subfigure}

  \caption{
  Examples from the RePo bottom test set. ViTPose-s estimates when trained on COCO (left) and on RePoGen data (right).
  Colors -- 
  \colorbox{yellow}{right hand}, \colorbox{orange}{right leg}, \colorbox{SkyBlue}{left hand} and  \colorbox{blue}{\textcolor{white}{left leg}}
  }
    \label{fig:RePo-bottom-test}

\end{figure}

\begin{figure}[tb]
  \centering
  \begin{subfigure}{\linewidth}
    \centering
    \includegraphics[width=\linewidth]{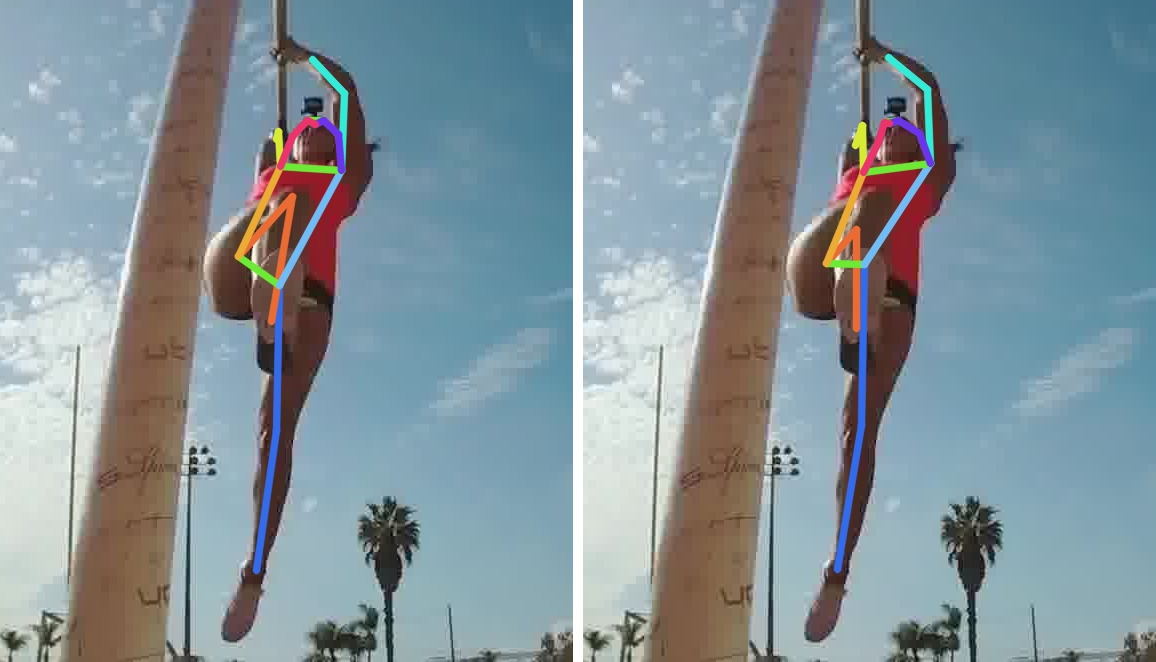}
  \end{subfigure}
  
  \begin{subfigure}{\linewidth}
    \centering
    \includegraphics[width=\linewidth]{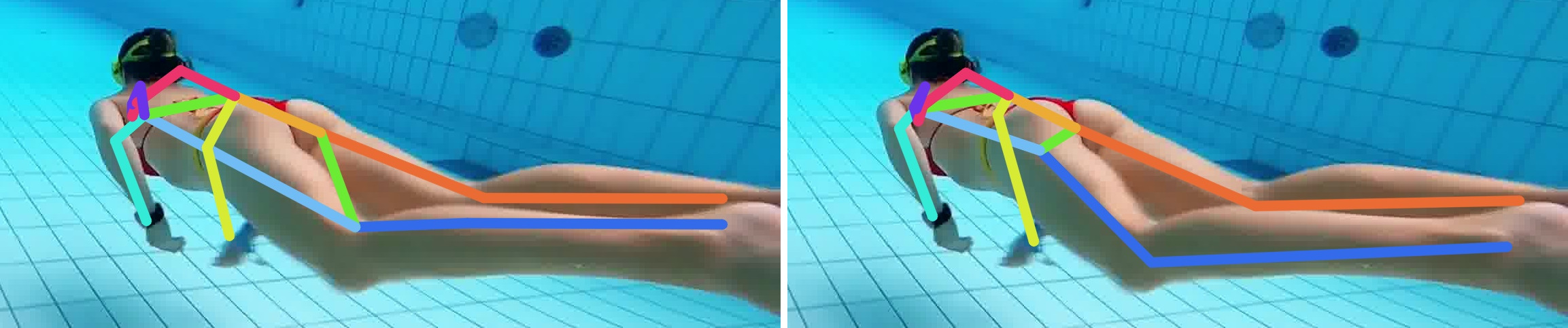}
  \end{subfigure}
  
  
  \begin{subfigure}{\linewidth}
    \centering
    \includegraphics[width=\linewidth]{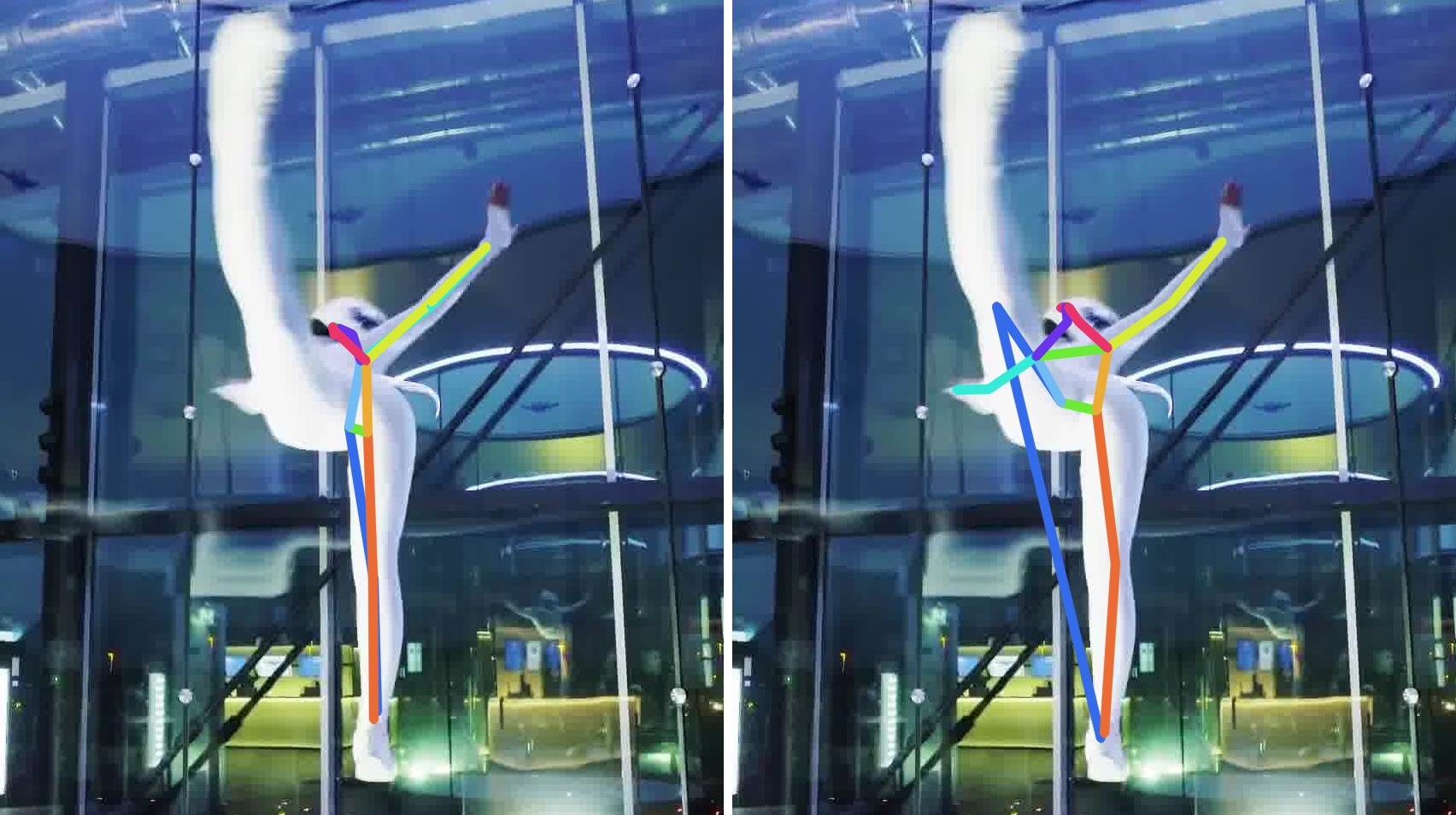}
  \end{subfigure}
  
  \begin{subfigure}{\linewidth}
    \centering
    \includegraphics[width=\linewidth]{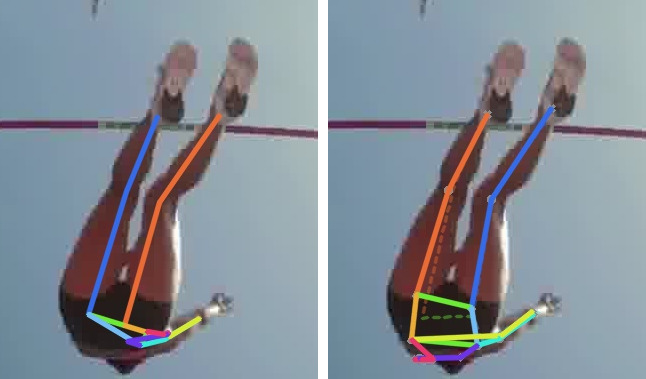}
  \end{subfigure}
  
  \begin{subfigure}{\linewidth}
    \centering
    \includegraphics[width=\linewidth]{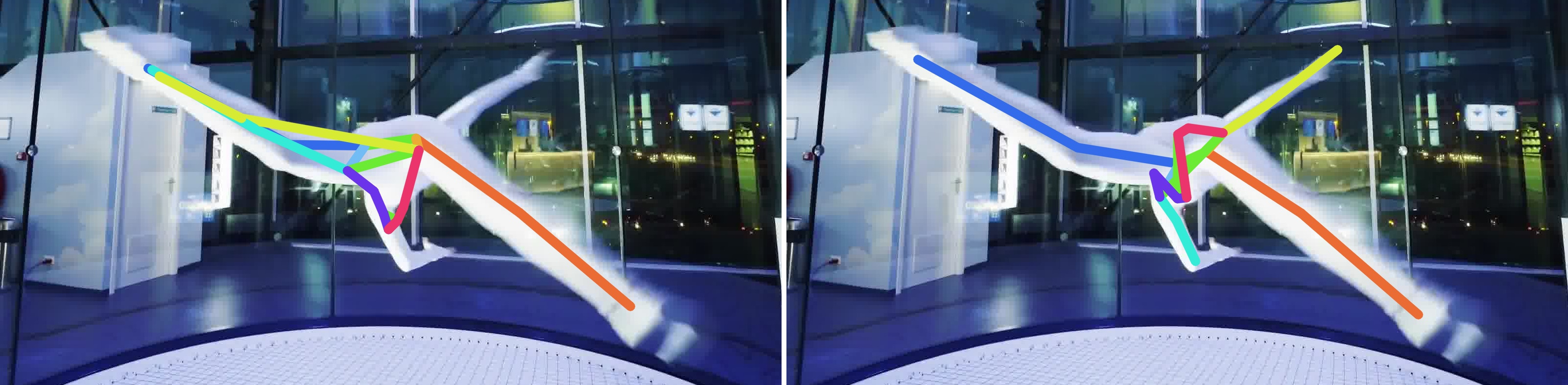}
  \end{subfigure}
  
  \caption{
  Examples from the RePo bottom val set. ViTPose-s estimates when trained on COCO (left) and on RePoGen data (right).
  Colors -- 
  \colorbox{yellow}{right hand}, \colorbox{orange}{right leg}, \colorbox{SkyBlue}{left hand} and  \colorbox{blue}{\textcolor{white}{left leg}}
  }
    \label{fig:RePo-bottom-val}

\end{figure}

\begin{figure}[tb]
  \centering
  \begin{subfigure}{\linewidth}
    \centering
    \includegraphics[width=0.95\linewidth]{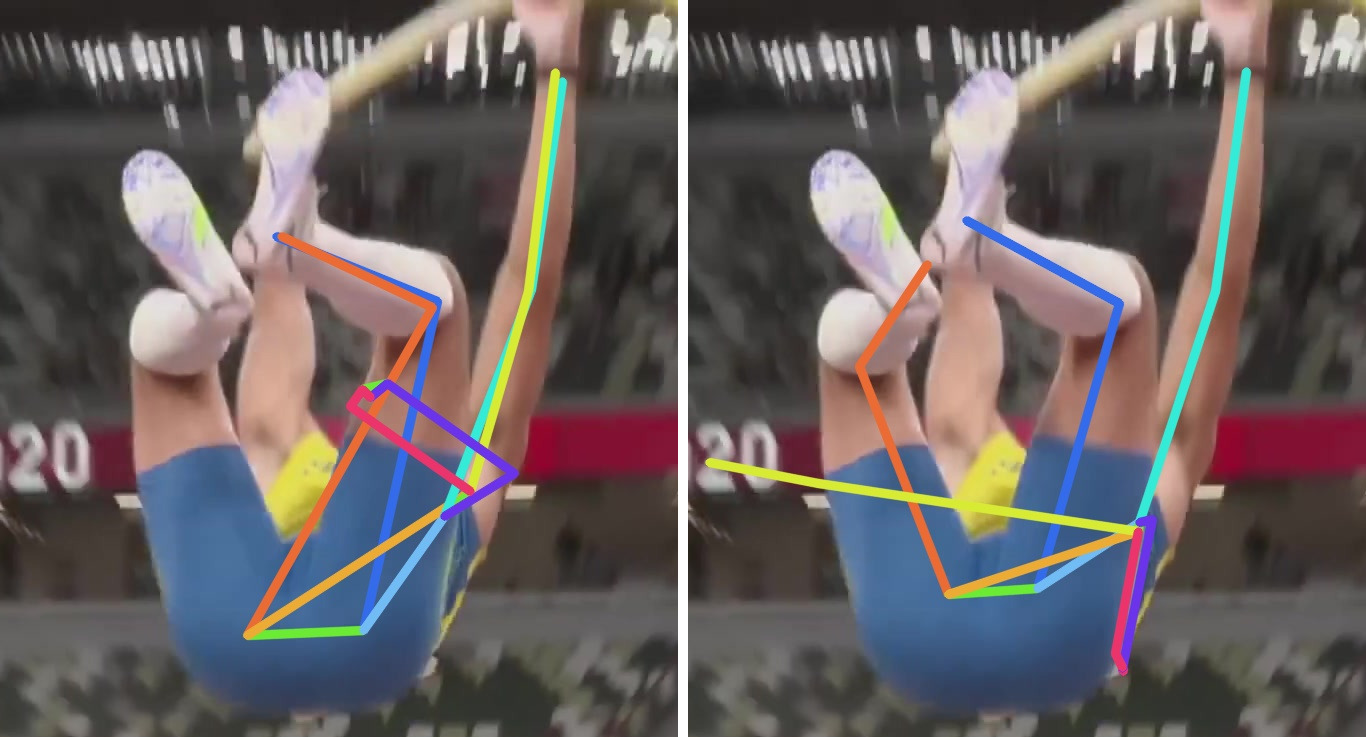}
  \end{subfigure}
  
  \begin{subfigure}{\linewidth}
    \centering
    \includegraphics[width=0.95\linewidth]{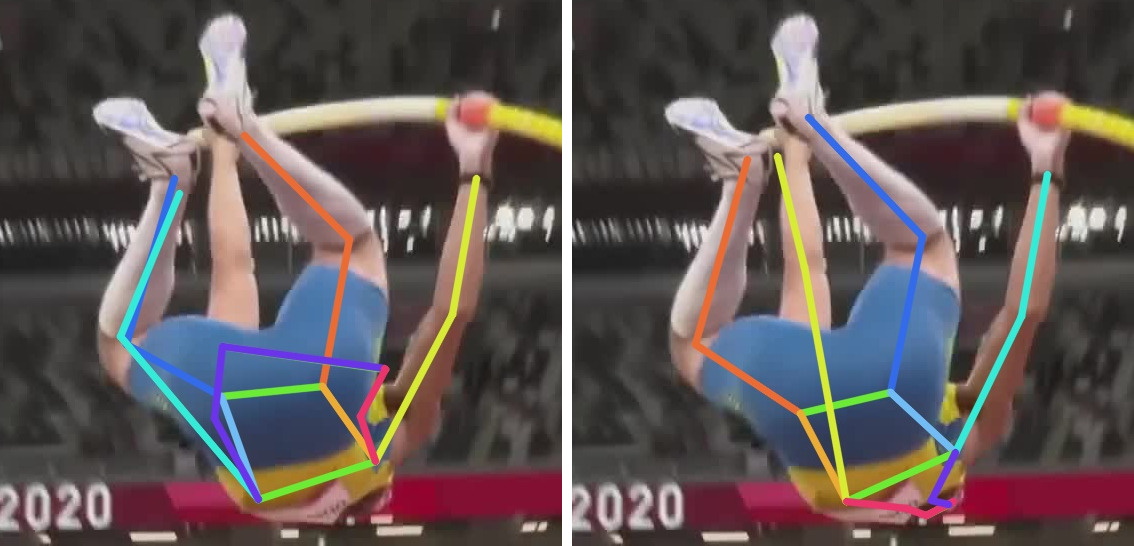}
  \end{subfigure}
  
  \begin{subfigure}{\linewidth}
    \centering
    \includegraphics[width=0.95\linewidth]{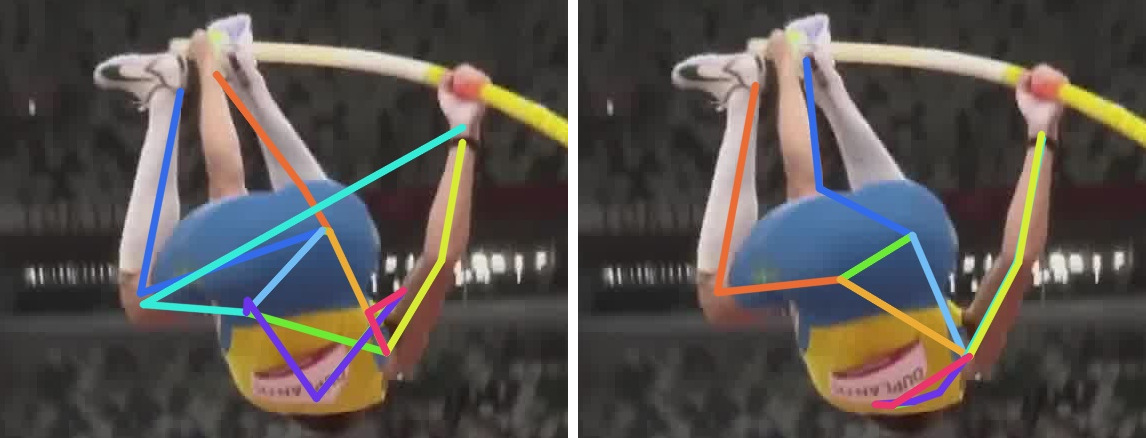}
  \end{subfigure}
  
  \begin{subfigure}{\linewidth}
    \centering
    \includegraphics[width=0.95\linewidth]{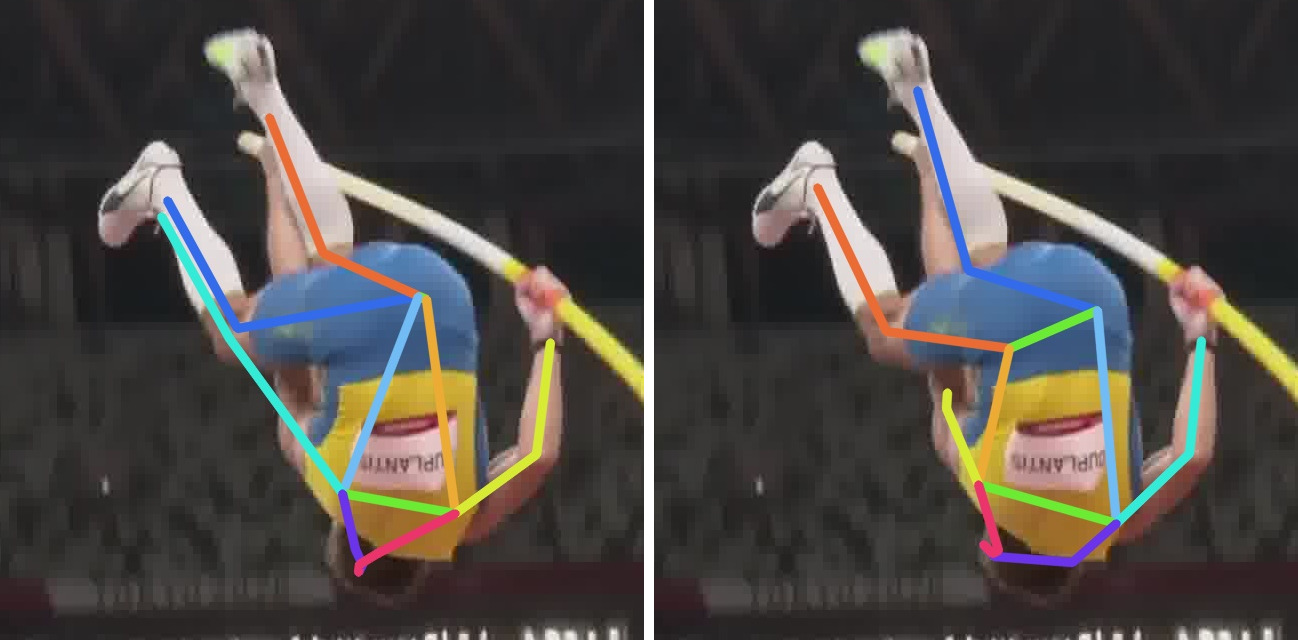}
  \end{subfigure}
  
  \begin{subfigure}{\linewidth}
    \centering
    \includegraphics[width=0.95\linewidth]{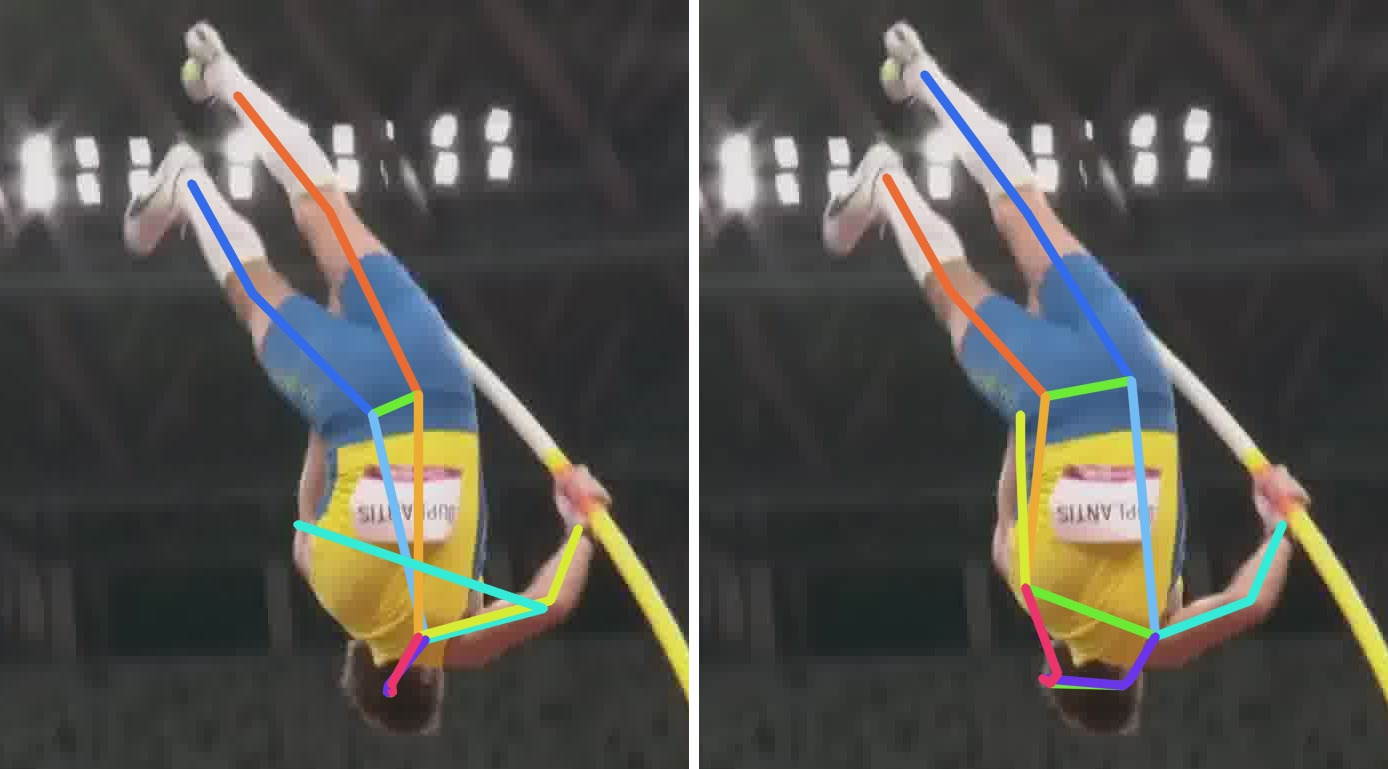}
  \end{subfigure}

  \caption{
  Examples from the RePo bottom seq set. ViTPose-s estimates when trained on COCO (left) and on RePoGen data (right). Images from a consecutive sequence, taking every third frame.
  Colors -- 
  \colorbox{yellow}{right hand}, \colorbox{orange}{right leg}, \colorbox{SkyBlue}{left hand} and  \colorbox{blue}{\textcolor{white}{left leg}}
  }
    \label{fig:RePo-bottom-seq}

\end{figure}

We also offer additional qualitative results on the proposed RePo dataset. The RePo Bottom Test is in the \cref{fig:RePo-bottom-test}, and Val set in \cref{fig:RePo-bottom-val}. The model fine-tuned with RePoGen data struggles the most with head keypoints and strong motion blur. The \cref{fig:RePo-bottom-seq} shows results on the Bottom Seq set where we show performance on a video. We show every third frame from a video.

\section{Code}
\label{sec:code}

The code is available in the supplementary material. The repository builds on the SMPL-X project \cite{SMPL-X} and uses the same dependencies. For more details, see the enclosed README.md file and the code itself.

\end{document}